\documentclass[10pt,twocolumn,letterpaper]{article}

\usepackage[pagenumbers]{arxiv} %
\usepackage{array}
\usepackage{textcomp}
\usepackage{stfloats}
\usepackage{url}
\usepackage{graphicx}
\usepackage{amsmath,amssymb,amsfonts}
\usepackage{multirow,multicol}
\usepackage{todonotes,xspace}
\usepackage[pagebackref,breaklinks,colorlinks,citecolor=cvprblue]{hyperref}
\usepackage{xcolor}
\definecolor{lightgraybg}{RGB}{245,245,245}
\definecolor{deepgrayborder}{RGB}{120,120,120}
\usepackage[table]{xcolor}
\definecolor{verylightpink}{HTML}{FFD6D6}
\definecolor{brightgreen}{RGB}{0,176,80}
\usepackage{tikz}

\usepackage{algorithm}
\usepackage{algpseudocode}
\usepackage{tcolorbox}      %
\tcbuselibrary{most} %
\usepackage{enumitem}       %
\usepackage{listings}
\usepackage{booktabs}
\usepackage{multirow}
\usepackage{graphicx} %
\usepackage{algorithm}
\usepackage{colortbl} 
\usepackage{pifont}
\usepackage{microtype}
\usepackage{overpic}

\definecolor{cvprblue}{rgb}{0.21,0.49,0.74}

\title{Obstruction reasoning for robotic grasping}

\author{
Runyu Jiao$^{1,2}$\qquad
Matteo Bortolon$^{1}$\qquad
Francesco Giuliari$^{1}$\qquad
Alice Fasoli$^{1}$\\
Sergio Povoli$^{1}$\qquad
Guofeng Mei$^{1}$\qquad
Yiming Wang$^{1}$\qquad
Fabio Poiesi$^{1}$\\[2mm]
$^{1}$Fondazione Bruno Kessler\quad
$^{2}$University of Trento\\[1mm]
{\tt\small \{rjiao, ywang, poiesi\}@fbk.eu}
}

\begin{document}
\newcommand{\fabiocomment}[1]{\todo[color=yellow!20, inline, author=Fabio]{#1}}
\newcommand{\sergio}[1]{\todo[color=green!20, inline, author=Sergio]{#1}}
\newcommand{\alice}[1]{\todo[color=red!20, inline, author=Alice]{#1}}
\newcommand{\runyu}[1]{\todo[color=olive!20, inline, author=Runyu]{#1}}
\newcommand{\yiming}[1]{\textcolor{red}{[YW: #1]}}
\newcommand{\francesco}[1]{\todo[color=orange!20, inline, author=Francesco]{#1}}
\newcommand{\guofeng}[1]{\todo[color=pink!20, inline, author=Guofeng]{#1}}
\newcommand{\matteo}[1]{\todo[color=blue!20, inline, author=Matteo]{#1}}

\newcommand{\higherbetter}[0]{{\color{black!50}{$\,\uparrow$}}}
\newcommand{\oracle}[1]{\textcolor{gray}{#1}}

\newcommand{\impp}[1]{{\textcolor{Green}{+#1}}}
\newcommand{\impn}[1]{{\textcolor{BrickRed}{-#1}}}

\newcommand{\benchmarkshort}{BENCHMARK\xspace}
\newcommand{\benchmarkfull}{BENCHMARK\xspace}

\newcommand{\ourmethod}{UNOGrasp\xspace}
\newcommand{\ourdataset}{UNOBench\xspace}

\newcommand{\suppmat}{\textit{Supp. Mat.}}

\newcommand{\cmark}{\ding{51}}
\newcommand{\xmark}{\ding{55}}

\definecolor{lightblue}{RGB}{203, 220, 235}
\definecolor{lightgreen}{RGB}{219,234,210}
\definecolor{lightred}{RGB}{255,130,130}
\definecolor{lightyellow}{RGB}{255, 254, 200}

\newcommand{\image}{I}
\newcommand{\imagedepth}{\image_{d}}
\newcommand{\imagergb}{\image_{rgb}}
\newcommand{\textspace}{\mathcal{L}}

\newcommand{\edit}[1]{\textbf{\color{red}#1}}

\newcommand{\inlineColorbox}[2]{\begingroup\setlength{\fboxsep}{1pt}\colorbox{#1}{\hspace*{2pt}\vphantom{Ay}#2\hspace*{2pt}}\endgroup}

\renewcommand{\arraystretch}{0.95}

\twocolumn[{%
    \renewcommand\twocolumn[1][]{#1}%
    \vspace{-22mm}
    \maketitle
    \thispagestyle{empty}
    \begin{center}
\includegraphics[width=\linewidth]{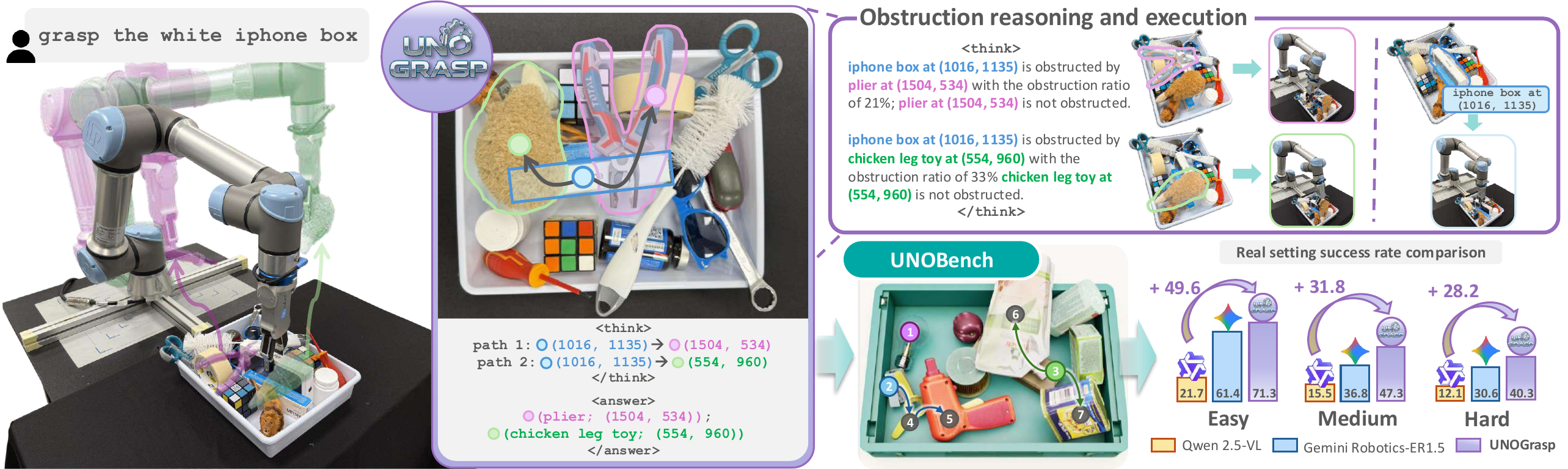}
\vspace{-6mm}
\captionof{figure}{\ourmethod performs multi-step obstruction reasoning for robotic grasping in cluttered scenes. 
Given an RGB-D image and a natural-language goal (e.g., \texttt{grasp the white iphone box}), \ourmethod reasons and grounds spatial information to infer the sequence of steps to unobstruct a requested object. 
We also introduce \ourdataset to comprehensively benchmark obstruction reasoning.
}
\label{fig:teaser}
\end{center}

}]

\begin{abstract}

Successful robotic grasping in cluttered environments not only requires a model to visually ground a target object but also to reason about obstructions that must be cleared beforehand. 
While current vision-language embodied reasoning models show emergent spatial understanding, they remain limited in terms of obstruction reasoning and accessibility planning.
To bridge this gap, we present \ourmethod, a learning-based vision-language model capable of performing visually-grounded obstruction reasoning to infer the sequence of actions needed to \underline{uno}bstruct the path and \underline{grasp} the target object.
We devise a novel multi-step reasoning process based on obstruction paths originated by the target object. 
We anchor each reasoning step with obstruction-aware visual cues to incentivize reasoning capability.
\ourmethod combines supervised and reinforcement finetuning through verifiable reasoning rewards.
Moreover, we construct \ourdataset, a large-scale dataset for both training and \underline{bench}marking, based on MetaGraspNetV2, with over 100k obstruction paths annotated by humans with obstruction ratios, contact points, and natural-language instructions. 
Extensive experiments and real-robot evaluations show that \ourmethod significantly improves obstruction reasoning and grasp success across both synthetic and real-world environments, outperforming generalist and proprietary alternatives. Project website: https://tev-fbk.github.io/UnoGrasp/.
\end{abstract}

\section{Introduction}\label{sec:intro}

Making robots interact with highly cluttered and unstructured 3D environments, such as bin-picking or object assembly following natural-language instructions, is an important skill for robotic manipulation~\cite{wang2025demonstrating}.
Successful grasping of a target object that is requested in natural language, demands Vision-Language Models (VLMs) \cite{bai2025qwen2} to not only visually ground and differentiate the target object, but also understand inter-object physical dependencies, particularly \textit{obstructions} among objects within the scene.
When objects impose on one another, the resulting physical obstruction can make manipulators fail in cluttered settings, as it prevents the robot's end-effector from successfully accessing the target object~\cite{jiao2025free}. 
While detection-based approaches can estimate obstruction relationships~\cite{Rabino2025graspplanning}, their design does not extend to broader embodied reasoning or multi-step action planning required by VLMs. 
Although VLM-based spatial reasoning is crucial for robotic manipulation, emerging benchmarks~\cite{du2024embspatial, song2025robospatial, wang2025spatial457, pothiraj2025capture} reveal that existing VLMs are generally limited in spatial reasoning necessary for physical interaction in the embodied context.
The challenge inherent to dense, cluttered scenes, where objects physically obstruct one another, still remains largely underexplored.

Preliminary research by Jiao \etal~\cite{jiao2025free} explores VLMs' zero-shot ability, exploiting Molmo~\cite{deitke2024molmopixmoopenweights} for visual grounding and prompting GPT-4o~\cite{achiam2023gpt} to reason whether to clear obstructing objects first. The recent release of (\textit{proprietary}) Gemini Robotics ER~\cite{abdolmaleki2025gemini} features a generalist model that exhibits interesting spatial understanding and grounding abilities.
While promising, current research~\cite{jiao2025free,abdolmaleki2025gemini} remains shallow in the task formalization and lacks in-depth investigation on how to evaluate and promote obstruction reasoning.

We advance embodied spatial reasoning for robotic grasping in clutter, primarily focusing on \textit{obstruction reasoning} with the objective of identifying obstruction paths directed from user-requested target objects.
We aim to \underline{bench}mark and enhance existing VLMs on obstruction reasoning capability, in order to \underline{uno}bstruct the paths and promote successful \underline{grasp} of the target object. 
To this end, we introduce \ourdataset, a dataset for both training and benchmarking VLMs' obstruction reasoning. 
\ourdataset is based on MetaGraspNetV2~\cite{Gilles2024MetaGraspNetV2}, featuring diverse daily objects in both synthetic and real scenes.
We associate each object with a human-annotated natural-language description to uniquely identify the object in clutter. 
\ourdataset provides 100k+ obstruction paths with rich metadata, \eg obstruction ratios, contact points, natural-language descriptions, which can be used to automatically generate multi-step obstruction reasoning process based on the obstruction paths.
We also propose a set of evaluation metrics to quantify models' reasoning performance at both object and obstruction-path levels.

Moreover, we introduce \ourmethod, a VLM equipped with novel visually-grounded obstruction reasoning ability for inferring the sequence of accessible obstructing objects that must be removed. 
\ourmethod addresses obstruction reasoning via formulating a directed graph with objects as nodes and obstruction relations as edges, allowing it to effectively infer accessible obstructors.
\ourmethod is trained on a portion of the \ourdataset synthetic dataset using a two-stage approach: supervised fine-tuning (SFT) to initialize its reasoning capability, then reinforcement fine-tuning (RFT) based on verifiable rewards and obstruction-aware visual cues to boost model's reasoning.
We benchmark \ourmethod against Gemini Robotics-ER 1.5~\cite{abdolmaleki2025gemini} and Qwen2.5-VL~\cite{bai2025qwen2} baselines. 
\ourmethod outperforms Gemini Robotics-ER 1.5 in both synthetic and real scenes of \ourdataset. 
Notably, we also conduct real-world robotic experiments in a laboratory environment (involving diverse objects and layouts), confirming \ourmethod's advantage over Gemini Robotics-ER 1.5 in terms of obstruction reasoning. 
\textit{Unlike proprietary models, we will release data, model and code publicly.}

\noindent In summary, our main contributions are:
\begin{itemize} 
\item We pioneer the deep study of spatial obstruction reasoning for robotic grasping in challenging cluttered scenes. 
\item We introduce \ourdataset, the first large-scale benchmark for training and testing obstruction reasoning, with evaluation protocols and metrics to quantify reasoning accuracy.
\item We propose \ourmethod, a VLM trained with a novel graph-based recipe that encourages obstruction reasoning with obstruction-aware visual cues, like occlusion ratio. 
\item \ourdataset confirms that obstruction reasoning remains an open challenge in embodied spatial reasoning, while \ourmethod achieves state-of-the-art performance.
\end{itemize}

\section{Related work}\label{sec:related}

\noindent\textbf{Spatial reasoning with VLMs.}
VLMs are limited in 3D spatial reasoning despite high VQA performance~\cite{wang2025spatial457, pothiraj2025capture}. 
Research addresses this by fine-tuning models with explicit 3D knowledge, such as metric distances (SpatialVLM~\cite{chen2024spatialvlm}) or depth inputs (SpatialBot~\cite{cai2025spatialbot}). 
Further progress utilizes reinforcement learning for robotic manipulation tasks~\cite{zhou2025roboreferspatialreferringreasoning,song2025maniplvm}, exploits intermediate representations \cite{yuan2025seeing} or descriptive scene graphs~\cite{qi2025sofar}, and employs visual prompting techniques (\eg, Set-of-Mark~\cite{yang2023set,yang2025magma}) to enhance reasoning. 
Affordance understanding is also being integrated (RoboPoint~\cite{yuan2024robopoint}, $A_0$~\cite{xu2025a0}), and VISO-Grasp~\cite{shi2025viso} tackles visibility constraints. 
Yet, most works on robotic manipulation do not account for scenarios where target objects are obstructed, thus hindering successful manipulation, while \ourmethod addresses this exact challenge.

\noindent\textbf{Obstruction reasoning in robotic grasping.} 
Grasping obstructed objects is challenging, requiring the robot to infer its occluded shape and reliable grasp locations from partial data~\cite{tziafas2023languageguided, liu2025fetchbot}. 
An equally important challenge is inferring the sequence of actions needed to clear complex arrangements (\eg, stacked objects) for access~\cite{zhang2018visual,Rabino2025graspplanning, jiao2025free}.
Recent VLM approaches for cluttered grasping include dedicated planners like RelationGrasp~\cite{liu2024relationgrasp} and GOAL~\cite{li2025graspocclusion}, and reasoning models like ThinkGrasp~\cite{qian2024thinkgrasp} and FreeGrasp~\cite{jiao2025free}, which use LLMs for object removal planning and visual prompting, respectively. 
Unlike prior methods that construct complex scene-level graphs covering all object pairs~\cite{Gilles2024MetaGraspNetV2, Rabino2025graspplanning}, \ourmethod~grounds the language to the target object and then builds a compact obstruction graph rooted exclusively at that target. 
\ourmethod focuses its reasoning on the structure dictating the target’s accessibility and grasp plan.

\noindent\textbf{Datasets and benchmarks on obstruction reasoning.}
Progress in obstruction reasoning include early works like VMRN~\cite{zhang2018visual} and REGRAD~\cite{zhang2022regard} that use object relationships to formulate obstruction graphs, while MetaGraspNetV2~\cite{Gilles2024MetaGraspNetV2} and amodal segmentation datasets like UOAIS-SIM~\cite{back2022unseen} focus on occlusion-related challenges. 
However, these datasets are not aligned with complex, reasoning-oriented manipulation tasks that require understanding multi-object obstruction chains. 
VLM benchmarks (\eg, EmbSpatial-Bench~\cite{du2024embspatial}, Spatial457~\cite{wang2025spatial457}, CAPTURe~\cite{pothiraj2025capture}) primarily test static perception rather than action-centric obstruction reasoning (i.e., planning clearance actions). 
However, most datasets lack language annotations, limiting their utility for VLMs and preventing the study of linguistically-grounded obstruction reasoning. 
To bridge this gap, we introduce \ourdataset, the first benchmark with annotated free-form language object descriptions to enable VLMs to jointly reason about occlusions and corresponding unobstructing actions.

\section{\ourdataset}\label{sec:benchmark}

\begin{figure*}[t]
    \centering
    \includegraphics[width=\linewidth]{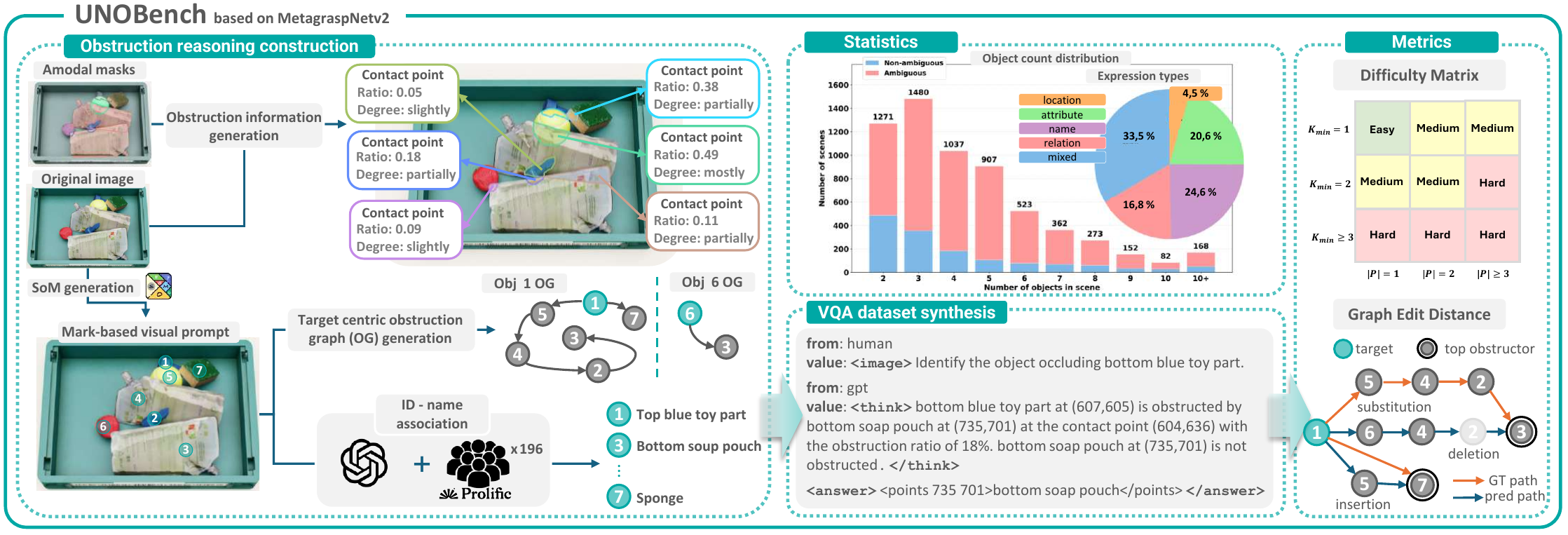}
    \vspace{-7mm}
    \caption{
    \ourdataset features two unique characteristics: (i) human-annotated free-form language instructions about objects in cluttered bins, and (ii) per-bin obstruction graphs for grounded spatial reasoning.
    Human annotators through the Prolific platform were involved to refine the initial GPT-4o generated annotations.
    \ourdataset features three levels of difficulty and introduces novel evaluation metrics.
    }
    \label{fig:dataset}
\end{figure*}

\ourdataset is built upon MetaGraspNetV2~\cite{Gilles2024MetaGraspNetV2} dataset, and is both for training and benchmarking (\cref{fig:dataset}). 
MetaGraspNetV2 provides amodal segmentation and object geometry, but it lacks explicit supervision for high-level reasoning and language grounding. 
\ourdataset introduces two unique characteristics: 
(i) human-annotated free-form language instructions of objects in cluttered bins, 
(ii) per-bin obstruction graphs for grounded spatial reasoning. 
\ourdataset enables obstruction-aware and language-guided grasping through, 
\textit{structured obstruction reasoning construction} to enrich each scene with physical obstruction information, object-centric graphs, and semantic knowledge, and
\textit{obstruction-aware VQA synthesis} to transform these structured annotations into a VQA dataset suitable for model training and evaluation.

\noindent\textbf{Structured obstruction reasoning construction.}
We semi-automatically construct a symbolic representation encoding visual obstruction structures in four steps:
(a) \emph{Set-of-Marks (SoM) preparation:} we overlay unique numeric marks \cite{yang2023set} on each object instance in the ground-truth masks, assigning an ID and centroid $(x, y)$;
(b) \emph{Obstruction information:} from amodal masks, we compute contact points, obstruction ratios, and obstruction degrees (slightly, partially, mostly, heavily obstructed);
(c) \emph{Object-centric obstruction graph:} for each target, we build a directed graph where nodes are SoM IDs and edges represent ``the obstructed $\rightarrow$ the obstructing'' relations with associated obstruction attributes;
(d) \emph{ID-name-coordinate association:} GPT-4o processes mark-based prompts to generate names for all IDs, forming $(\text{\textit{id}}, \text{\textit{name}}, (x,y))$ triplets.
We rely on human annotators to refine 5,400 challenging images, where 196 native speakers on Prolific reviewed 41,193 object names (80 minutes per person), followed by expert rechecking of all scenes, resulting in 4,678 corrected images and 17,261 revised object names, ensuring linguistic accuracy and visual consistency.

\noindent\textbf{Obstruction-aware VQA synthesis.}
We generate two complementary datasets following the structured \texttt{<think>} and \texttt{<answer>} format. 
These two datasets form a unified benchmark: the \textit{Oracle with Set-of-Mark (SoM)} dataset assesses structured reasoning with explicit grounding, while the \textit{Natural Language Prompting} dataset evaluates obstruction reasoning and grounding based on free-form instructions.
Specifically:
(i) Oracle (SoM): Starting from the object-centric obstruction graph (OG), we use predefined templates to generate questions and reasoning traces. 
All instances are represented by numeric IDs only (no names or coordinates).
This setting solely evaluates the model's reasoning capability, as all object instances are unambiguously identified via SoM.
(ii) Natural Language Prompting: Building upon the Oracle formulation, this setting better reflects real-world robot usage, where users' prompts are questions given in free-form language without explicit IDs or coordinates. 
The model-generated \texttt{<think>} and \texttt{<answer>} traces include both object names and coordinates at each reasoning step, reflecting realistic human–robot interaction. 
This dataset measures a model’s ability on both obstruction reasoning and spatial grounding with linguistic instructions.

\noindent\textbf{Dataset statistics.}
\ourdataset comprises synthetic and real scenes.
The former includes 6,255 scenes with 25,020 view images, 97,066 object instances annotated with names, and 108,174 reasoning paths.
The latter includes 520 scenes with one view per scene, 2,232 object instances annotated with names, and 2,552 obstruction paths.

\subsection{Evaluation protocol}\label{sec:eval_procedure}

We introduce our metrics and benchmark split below.

\noindent\textbf{Outcome-level metrics.}
For each target object, we quantify the correctness of the final answer (the predicted top-level obstruction set, $\mathcal{F}_{\text{pred}}(o_t)$) against the ground truth ($\mathcal{F}_{\text{GT}}(o_t)$). 
We report Success Rate Precision (SR-P), Recall (SR-R), and the F1-score, which collectively measure the accuracy of the model's final action output in the $\texttt{<answer>}$ section.

\noindent\textbf{Reasoning-level metrics.}
We assess the model’s reasoning ability in \texttt{<think>} at two levels.
\textit{Object-level reasoning} is computed using Object Triplet Precision (OP), Recall (OR), and $\text{F1}_{\text{rel}}$, as in~\cite{Rabino2025graspplanning}. 
For each pair of objects, we deem a true positive when both objects and their obstruction relationship are correctly identified.
\textit{Path-level reasoning} is computed using our new metric Multi-Path Normalized Edit Distance (MP\_NED).
MP\_NED measures the structural alignment between predicted ($\mathcal{P}$=$\{p_i\}_{i=1}^m$) and ground-truth ($\mathcal{G}$=$\{g_j\}_{j=1}^n$) reasoning paths.
Formally, $\text{NED}(p_i, g_j) = \frac{\text{EditDist}(p_i, g_j)}{\max(|p_i|, |g_j|)}$, where EditDist is the Levenshtein distance.
We then find the minimal-cost assignment via the Hungarian algorithm using $C_{ij} = \text{NED}(p_i, g_j)$ as the cost matrix. 
MP\_NED is the mean cost over matched pairs: $\text{MP\_NED} = \max(m,n)^{-1} \sum_{(i,j)\in \text{match}} C_{ij}$. 
A lower MP\_NED indicates closer structural alignment of the reasoning paths.
More details are referred to \suppmat

\noindent\textbf{Difficulty-based evaluation split.}
We categorize the difficulty of each target object based on its obstruction graph depth ($K_{\min}$) and the number of distinct reasoning paths ($|P|$). 
We divide the benchmark into four difficulty levels (\textit{No-Occ}, \textit{Easy}, \textit{Medium}, and \textit{Hard}), as summarized below:
\begin{table}[h]
\tabcolsep 4pt
\vspace{-2mm}
\resizebox{\columnwidth}{!}{%
\begin{tabular}{lll}
\toprule
\textbf{Level} & \textbf{Condition} & \textbf{Interpretation} \\
\midrule
\cellcolor{gray!20}No-Occ & $K_{\min}=0$ & No obstruction \\[2pt]
\cellcolor{lightgreen}Easy & $K_{\min}=1,\, |P|=1$ & Single-path reasoning \\[2pt]
\cellcolor{lightyellow}
& ($K_{\min}=1,\, |P|>1$) or 
&  \\[2pt]
\multirow{-2}{*}{\cellcolor{lightyellow}Medium} & ($K_{\min}=2,\, |P|\le2$) & \multirow{-2}{*}{\centering Multi-path or shallow depth}\\[2pt]
\cellcolor{verylightpink}Hard & $K_{\min}\ge3$ or ($K_{\min}=2,\, |P|>2$) & Deep or complex reasoning \\
\bottomrule
\end{tabular}
}
\end{table}

\section{\ourmethod}
\label{sec:method}

\noindent\textbf{Problem formulation.}
Let us define a cluttered workspace with a set of $N$ visible objects as $\mathcal{O} = \{o_1, o_2, \ldots, o_N\}$ (\cref{fig:dataset}).
Given an RGB-D observation $\image = (\imagergb, \imagedepth)$ and a free-form textual instruction $q\in\textspace$ (e.g., \texttt{grasp the white box on the leftmost}) that uniquely refers to a target object $o_t\in\mathcal{O}$, we aim to produce an unobstruction plan to grasp $o_t$. 
$\imagergb$ is used for reasoning and action sequence planning. 
$\imagedepth$ is used to estimate 3D grasping points. 
If $o_t$ is unobstructed, \ourmethod instructs a direct grasp. 
Otherwise, \ourmethod identifies the minimal sequence of actions to access $o_t$. 
This sequence begins by identifying all top-level obstructing objects, those that obstruct $o_t$ but are themselves accessible, \ie free of obstruction. Each action corresponds to an object removal.
Note that our action plan is purely based on the obstruction's existence.
Various obstructions may impede grasping differently, thus quantifying obstruction severity is highly challenging and application-dependent, out of the scope of this paper.

\begin{figure*}[t]
    \centering
    \includegraphics[width=\linewidth]{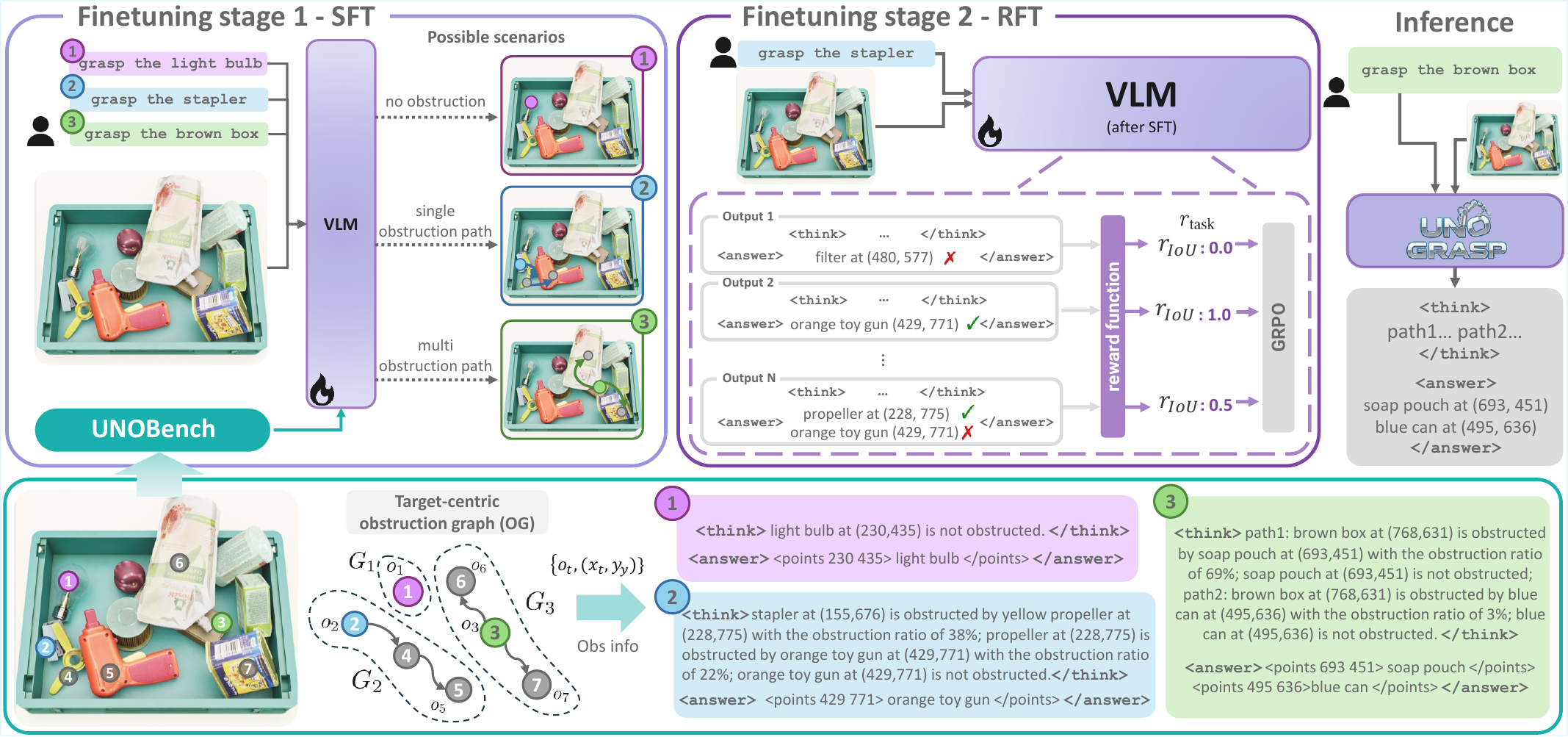}
    \vspace{-7mm}
    \caption{
    \ourmethod is a VLM trained through supervised fine (SFT) on \ourdataset to learn structured obstruction-path reasoning, and through GRPO-based reinforcement finetuning (RFT) to further boost its reasoning ability using outcome-driven IoU and format rewards. 
    During inference, given an RGB image and a target object as language instruction, \ourmethod reasons over multiple obstruction paths ($\texttt{<think>}$ traces) and directly outputs the sequence of actions ($\texttt{<answer>}$) required to remove obstructions and grasp the target.
    }
    \label{fig:method}
\end{figure*}

\ourmethod formulates obstruction reasoning as a directed graph.
We train it on a portion of \ourdataset's synthetic set using a two-stage approach: supervised fine-tuning (SFT) to initialize its reasoning capability, then reinforcement fine-tuning (RFT) based on obstruction-aware visual cues to boost model's reasoning.
Fig.~\ref{fig:method} illustrates \ourmethod stages.

\subsection{Target-centric obstruction graph}\label{sec:graph}

Given the instruction $q$ in free-form language, the model performs spatial reasoning to ground the linguistic reference in $\imagergb$ and identify $o_t$ among all visible objects.
If $o_t$ is not initially visible (\eg deeply beneath), the model uses contextual cues to infer the most plausible candidate.
Once $o_t$ is localized, we model its visual obstruction. 
Instead of reasoning over all possible pairwise obstruction relations~\cite{Rabino2025graspplanning}, we construct a target-centric obstruction graph that exclusively captures the objects relevant to the accessibility of $o_t$.

The sequence of actions required to unobstruct the target object $o_t$ can be modeled as a directed graph, $G_t = (\mathcal{V}_t, E_t)$. 
The node set $\mathcal{V}_t$ includes the target object $o_t$ and all objects that directly or indirectly obstruct it. 
A directed edge $(o_i, o_j) \in E_t$ indicates that object $o_i$ is obstructed by object $o_j$ when viewed from the camera viewpoint. 
Edges are directed from the obstructed object to the obstructors, forming one or more obstruction paths that originate at the target object $o_t$ and terminate at the accessible top-level obstructors.
Any object that appears along one of these obstruction paths can be considered an ancestor of $o_t$, and must be removed before $o_t$ becomes fully accessible.
Let $\mathcal{A}(o_t)$ be the set of ancestor objects of $o_t$ that is defined as
\begin{equation}
\mathcal{A}(o_t) {=}
\{\, o_i {\in} \mathcal{O} \mid 
\exists \text{ a directed path } [o_t, {\cdots}, o_i] 
{\in} G_t \,\}.
\end{equation}
Objects that lie on top of the clutter are defined as
\begin{equation}
\mathcal{F}(o_t) {=}
\{\, o_i {\in} \mathcal{A}(o_t) 
\mid \nexists\, o_j \text{ s.t. } (o_i, o_j) {\in} E_t \,\}.
\end{equation}
Each $o_i \in \mathcal{F}(o_t)$ is a visible and graspable object; removing any of these will reduce the obstruction of $o_t$.

Lastly, we express the reasoning objective as:
\begin{equation}\label{eq:function_theta}
f_{\Theta}(I, q) =
\begin{cases}
\mathcal{A}(o_t) \rightarrow \mathcal{F}(o_t), & \text{if } o_t \text{ is obstructed,}\\
o_t, & \text{otherwise.}
\end{cases}
\end{equation}
$f_{\Theta}$ is parametric, with $\Theta$ being its parameters. We aim to train $f_{\Theta}$ to output $\mathcal{F}(o_t)$ through obstruction reasoning $\mathcal{A}(o_t)$.
When multiple top-level obstructors exist,  
$\mathcal{F}(o_t)$ provides a set of next-step candidates,  
from which the robot can select based on constraints like graspability, or reachability.

$\text{\cref{fig:method}}$ shows three examples of target-centric obstruction graphs: $G_1$, $G_2$, and $G_3$.
For the light bulb ($o_1$), which is unobstructed, its graph $G_1$ contains only the object itself. 
$G_2$ represents the obstruction graph of the stapler ($o_2$). 
The stapler is blocked by the yellow propeller ($o_4$), which is further blocked by the orange toy gun ($o_5$). 
Thus, both objects are ancestors, $\mathcal{A}(o_2)=\{o_4, o_5\}$. 
However, only the orange toy gun is itself unobstructed, making it the top-level obstructing set: $\mathcal{F}(o_2)=\{o_5\}$. 
In $G_3$, both the soap pouch ($o_6$) and blue can ($o_7$) obstruct the brown box ($o_3$) and are themselves unobstructed, meaning $\mathcal{F}(o_3)=\mathcal{A}(o_3)=\{o_6, o_7\}$.
Ultimately, the model’s objective is to accurately infer these minimal obstruction sets, $\mathcal{F}(o_t)$, which guide the robot in deciding the next necessary action. 
We next describe how $f_{\Theta}$ is trained to achieve this reasoning capability.

\subsection{Training pipeline}
We develop a two-stage training pipeline to train $f_{\Theta}$ (Eq.~\ref{eq:function_theta}) with visually grounded, obstruction-aware reasoning:
i) warm-start supervised finetuning (SFT) that encourages the model to output visually grounded reasoning chains aligned with the obstruction graph of the referred target;
ii) reinforcement finetuning (RFT) that optimizes task-relevant behaviors using our novel obstruction-aware rewards.

\noindent\textbf{SFT with visually-grounded chains.}
We finetune $f_{\Theta}$ on \ourdataset (\S\ref{sec:benchmark}) to interpret free-form language instructions ($q$) and associate them with a unique target object ($o_t$) in the visual scene. 
This grounding is supervised using two methods for explicit reference: the object's name and its image coordinates $\{ o_t, (x_t, y_t) \}$, and a SoM visual prompt, where objects are assigned unique IDs. 
These explicit cues are essential for disambiguating multiple instances based on spatial or relational cues in $q$.
The fine-tuning instructs $f_{\Theta}$ to identify if 
(i) $o_t$ is unobstructed, 
(ii) $o_t$ has a single obstruction path, and 
(iii) $o_t$ has multiple obstruction paths.
The model generates a step-by-step reasoning chain where every step is anchored to a physically adjacent (contacting) neighbor, ensuring the chain traverses valid obstructions. 
We encourage $f_{\Theta}$ to quantify obstruction levels as auxiliary signals within the chain, aligning with findings that spatial grounding encourages visual reasoning~\cite{sarch2025groundedreinforcementlearningvisual}. This process strengthens the model's ability to reconstruct complete obstruction paths and identify the top-level obstructors $\mathcal{F}(o_t)$ that constitute the next-step action set.

\noindent\textbf{RFT with obstruction-aware rewards.}
Starting from the SFT-bootstrapped model $f_{\Theta}$, we perform RFT to enhance its grounded reasoning ability. 
The model produces a natural language description and corresponding image coordinates for each object, enabling $f_{\Theta}$ to progressively refine its attention using task-relevant visual information~\cite{sarch2025groundedreinforcementlearningvisual}.
We adopt Group Relative Policy Optimization (GRPO)~\cite{shao2024deepseekmath} to average rewards across multiple sampled outputs, using a standard formulation similar to related works~\cite{sarch2025groundedreinforcementlearningvisual, zhou2025roboreferspatialreferringreasoning}.
Importantly, we define a novel, task-specific reward $r$ as a weighted combination of a format reward $r_{\text{fmt}}$ and a task reward $r_{\text{task}}$:
\begin{equation}
r = \lambda_{\text{fmt}} \, r_{\text{fmt}} + \lambda_{\text{task}} \, r_{\text{task}}.
\end{equation}
The format reward, $r_{\text{fmt}}$, is binary ($1$ or $0$), promoting structural validity by checking for the correct presence and closure of the reasoning $\texttt{<think>}$ and action $\texttt{<answer>}$ contexts. 
The task reward, $r_{\text{task}}$, supervises the grounded output $\mathcal{F}(o_t)$ (the content of the action context) using a set-level (Intersection over Union) IoU metric:
\begin{equation}
r_{\text{task}} = \frac{|\mathcal{F}_{\text{pred}}(o_t) \cap \mathcal{F}_{\text{gt}}(o_t)|} {|\mathcal{F}_{\text{pred}}(o_t) \cup \mathcal{F}_{\text{gt}}(o_t)|}.
\end{equation}
This IoU provides a smoother optimization signal than binary correctness, rewarding partially correct predictions for more stable learning. 
Although this reward only supervises the final prediction $\mathcal{F}(o_t)$, experiments in $\S\text{\ref{sec:ablations}}$ show it also contributes to improving the quality of the internal obstruction reasoning path $\mathcal{A}(o_t)$. 
For completeness, a path-level fidelity metric $r_{\text{path}}$ is also evaluated post hoc.

\section{Experiments}\label{sec:exps}

\begin{table*}[t]
\caption{Path-level reasoning results on \ourdataset synthetic test set.
ICL: In-Context Learning.
SFT: Supervised finetuning.
SR: Success Rate;
MP\_NED: Multi-Path Normalized Edit Distance;
P: Precision;
R: Recall;
F1: F1-Score;
Best results \textbf{bold}; 
second best \underline{underlined}.
}
\vspace{-6pt}
\tabcolsep 2pt
\label{tab:syn_metric1}
\resizebox{\textwidth}{!}{
\begin{tabular}{l cc |cccc| cccc |cccc}
\toprule
\multirow{2}{*}{Method} &
\multicolumn{2}{c}{\cellcolor{gray!20}No obstructions} &
\multicolumn{4}{c}{\cellcolor{lightgreen}Easy} &
\multicolumn{4}{c}{\cellcolor{lightyellow}Medium} &
\multicolumn{4}{c}{\cellcolor{verylightpink}Hard} \\
\cmidrule(lr){2-3} \cmidrule(lr){4-7} \cmidrule(lr){8-11} \cmidrule(lr){12-15}
& SR (\%)$\uparrow$ & MP\_NED$\downarrow$
& SR-P (\%)$\uparrow$ & SR-R (\%)$\uparrow$ & SR-F1 (\%)$\uparrow$ & MP\_NED$\downarrow$
& SR-P (\%)$\uparrow$ & SR-R (\%)$\uparrow$ & SR-F1 (\%)$\uparrow$ & MP\_NED$\downarrow$
& SR-P (\%)$\uparrow$ & SR-R (\%)$\uparrow$ & SR-F1 (\%)$\uparrow$ & MP\_NED$\downarrow$ \\
\midrule

\rowcolor{gray!15}
\multicolumn{15}{c}{Oracle (with SoM)} \\

Gemini Robotics-ER 1.5~\cite{abdolmaleki2025gemini}
& 68.7 & \underline{0.17}
& 57.7 & 62.8 & 59.3 & 0.25
& 33.0 & 30.2 & 29.8 & 0.56
& 5.9 & 5.7 & 5.4 & 0.74 \\

Gemini Robotics-ER 1.5~\cite{abdolmaleki2025gemini} (ICL)
& 54.6 & 0.25
& 67.1 & \underline{73.3} & 69.1 & \underline{0.20}
& 41.8 & 39.7 & 38.7 & \underline{0.50}
& 14.9 & 14.8 & 13.8 & \underline{0.68} \\

Qwen2.5-VL~\cite{bai2025qwen2} (ICL)
& 9.8 & 0.64
& 24.7 & 36.7 & 27.2 & 0.55
& 20.7 & 27.2 & 20.3 & 0.73
& 9.7 & 17.2 & 10.2 & 0.79 \\

Qwen2.5-VL~\cite{bai2025qwen2} (SFT)
& \underline{88.7}  & -
& \underline{69.6} & 70.3 & \underline{69.8} & -
& \underline{64.2} & \underline{53.1} & \underline{56.5} & -
& \underline{38.1} & \underline{33.3} & \underline{34.3} & - \\

\ourmethod 
& \textbf{94.8} & \textbf{0.03}
& \textbf{82.8} & \textbf{84.4} & \textbf{83.3} & \textbf{0.11}
& \textbf{74.8} & \textbf{67.2} & \textbf{69.1} & \textbf{0.37}
& \textbf{56.8} & \textbf{55.3} & \textbf{54.5} & \textbf{0.51} \\

\rowcolor{gray!15}
\multicolumn{15}{c}{Natural Language Prompting} \\

Gemini Robotics-ER 1.5~\cite{abdolmaleki2025gemini}
& 50.2 & 0.88 
& 51.8 & 52.8 & 52.1 & 0.84 
& 36.9 & 30.9 & 32.5 & 0.87 
& 11.7 & 9.4 & 10.1 & 0.91 \\

Gemini Robotics-ER 1.5~\cite{abdolmaleki2025gemini} (ICL)
& 45.3 & \underline{0.83} 
& 60.6 & 61.8 & 61.0 & \underline{0.80} 
& 45.5 & 37.3 & 39.5 & \underline{0.85} 
& 17.2 & 13.4 & 14.6 & \underline{0.89} \\

Qwen2.5-VL~\cite{bai2025qwen2} (ICL)
& 11.2 & 0.84 
& 11.8 & 13.1 & 12.2 & 0.86 
& 11.8 & 10.8 & 10.6 & 0.88 
& 9.7 & 9.4 & 8.9 & \underline{0.89} \\

Qwen2.5-VL~\cite{bai2025qwen2} (SFT)
& \underline{91.4} & - 
& \underline{65.2} & \underline{65.5} & \underline{65.3} & - 
& \underline{59.6} & \underline{47.9} & \underline{51.5} & - 
& \underline{33.9} & \underline{31.5} & \underline{31.9} & - \\

\ourmethod
& \textbf{92.5} & \textbf{0.06} 
& \textbf{74.8} & \textbf{75.1} & \textbf{74.9} & \textbf{0.20} 
& \textbf{68.0} & \textbf{55.8} & \textbf{59.7} & \textbf{0.53} 
& \textbf{42.2} & \textbf{35.4} & \textbf{37.2} & \textbf{0.67}   \\

\bottomrule
\end{tabular}
}
\end{table*}

\begin{table*}[t]
\caption{
Object-level reasoning results on \ourdataset synthetic test set.
ICL: In-Context Learning;
SFT: Supervised fine tuning;
OP: Object triplet Precision;
OR: Object triplet Recall;
F1\textsubscript{rel}: Object triplet F1-Score;
Best results \textbf{bold}; 
second best \underline{underlined}.
}
\vspace{-6pt}
\tabcolsep 14pt
\centering
\label{tab:syn_metric2}
\resizebox{\textwidth}{!}{
\begin{tabular}{lccc| ccc |ccc |ccc}
\toprule
\multirow{2}{*}{Method} &
\multicolumn{3}{c}{\cellcolor{lightgreen}Easy} &
\multicolumn{3}{c}{\cellcolor{lightyellow}Medium} &
\multicolumn{3}{c}{\cellcolor{verylightpink}Hard} &
\multicolumn{3}{c}{Overall} \\
\cmidrule(lr){2-4} \cmidrule(lr){5-7} \cmidrule(lr){8-10} \cmidrule(lr){11-13}
& OP$\uparrow$ & OR$\uparrow$ & F1\textsubscript{rel}$\uparrow$
& OP$\uparrow$ & OR$\uparrow$ & F1\textsubscript{rel}$\uparrow$
& OP$\uparrow$ & OR$\uparrow$ & F1\textsubscript{rel}$\uparrow$
& OP$\uparrow$ & OR$\uparrow$ & F1\textsubscript{rel}$\uparrow$ \\
\midrule

\rowcolor{gray!15}
\multicolumn{13}{c}{Oracle (with SoM)} \\

Gemini Robotics-ER 1.5~\cite{abdolmaleki2025gemini}
& 56.0 & 62.3 & 57.9 
& 57.0 & 30.6 & 37.4 
& 46.1 & 14.4 & 20.8 
& 56.0 & 51.2 & 50.6 \\

Gemini Robotics-ER 1.5~\cite{abdolmaleki2025gemini} (ICL)
& \underline{67.8} & \underline{76.5} & \underline{70.4} 
& \underline{69.1} & \underline{39.8} & \underline{47.3} 
& \textbf{67.6} & \underline{24.3} & \underline{33.0} 
& \underline{68.2} & \underline{63.8} & \underline{62.3} \\

Qwen2.5-VL~\cite{bai2025qwen2} (ICL)
& 19.6 & 33.3 & 22.8 
& 21.4 & 19.7 & 18.5 
& 13.9 & 10.8 & 10.6 
& 20.0 & 28.5 & 21.1 \\

\ourmethod (ours)
& \textbf{81.3} & \textbf{85.9} & \textbf{82.6}
& \textbf{77.8} & \textbf{57.3} & \textbf{62.0} 
& \underline{63.0} & \textbf{43.6} & \textbf{48.7} 
& \textbf{79.7} & \textbf{76.0} & \textbf{75.3} \\

\rowcolor{gray!15}
\multicolumn{13}{c}{Natural Language Prompting} \\

Gemini Robotics-ER 1.5~\cite{abdolmaleki2025gemini}
& 2.5 & 2.7 & 2.6 
& 4.6 & 2.1 & 2.8 
& \underline{7.8} & 2.0 & \underline{3.1} 
& 3.3 & 2.5 & 2.6 \\

Gemini Robotics-ER 1.5~\cite{abdolmaleki2025gemini} (ICL)
& \underline{3.4} & \underline{3.7} & \underline{3.5} 
& \underline{6.2} & 2.6 & \underline{3.5} 
& 6.1 & 1.7 & 2.6 
& \underline{4.3} & \underline{3.3} & \underline{3.5} \\

Qwen2.5-VL~\cite{bai2025qwen2} (ICL)
& 2.1 & 3.6 & 2.5 
& 4.0 & \underline{2.9} & 3.1 
& 3.7 & \underline{2.2} & 2.7 
& 2.7 & \underline{3.3} & 2.7 \\

\ourmethod (ours)
& \textbf{65.6} & \textbf{67.4} & \textbf{66.1} 
& \textbf{57.4} & \textbf{33.0} & \textbf{39.5} 
& \textbf{44.6} & \textbf{20.1} & \textbf{25.7} 
& \textbf{62.6} & \textbf{56.0} & \textbf{57.2}\\

\bottomrule
\end{tabular}
}
\end{table*}

We compare \ourmethod against two VLMs. 
We use Gemini Robotics-ER 1.5~\cite{abdolmaleki2025gemini} as proprietary baseline in two variants: 
base model, provided with a prompt and real output examples, and In-Context Learning (ICL), prompted with three few-shot examples covering all obstruction types (no, single-path, and multi-path obstruction). 
Coordinate expressions are adapted to the model's native syntax for both. 
We use Qwen2.5-VL-3B~\cite{bai2025qwen2} as open-source baseline in two variants: 
ICL, prompted as for Gemini ICL; 
SFT, finetuned on \ourdataset using the same supervised setup as \ourmethod but without the \texttt{<think>} reasoning part; 
\ourmethod is built on Qwen2.5-VL-3B, and trained with SFT and RFT.
We evaluate all methods on both the \textit{oracle (SoM)} and \textit{natural language prompting} splits of \ourdataset.
The synthetic scenes are split into training, validation, and testing sets with a 7:1:2 ratio. 
All the real scenes are exclusively used for testing. 
We follow the procedure detailed in \S\ref{sec:eval_procedure}.

\noindent\textbf{Implementation details.}
We train $f_\Theta$ on 4 A100-SXM-64GB GPUs, using 2 epochs for SFT, and 1 epoch for RFT (with a generation group size of 4). 
For Gemini Robotics-ER 1.5, temperature is set to 0.1, and thinking budget to 2000.

\subsection{Quantitative analysis}

\cref{tab:syn_metric1} and \ref{tab:syn_metric2} report path-level and object-level obstruction reasoning results, respectively, on the \ourdataset synthetic test set.
Similarly, \cref{tab:real_metric1} and \ref{tab:real_metric2} report these results on the \ourdataset real set.
\textit{UNOBench enhances both reasoning and decision accuracy.}
Models finetuned on UNOBench (Qwen2.5-VL (SFT) and \ourmethod) achieve important gains in recognizing top obstructors compared to Qwen2.5-VL (ICL). 
\ourmethod improvement in reasoning quality is larger, and surpasses the proprietary Gemini Robotics-ER 1.5 across most settings, even in real subsets that are never seen during training.
\textit{Reasoning ability is crucial for complex scenes.} 
As the obstruction level increases, the advantage of reasoning supervision grows. 
On the synthetic hard split, \ourmethod surpasses Qwen2.5-VL (SFT) by +20.2\% SR-F1; on the real hard split, the margin widens to +38.0\%, confirming that process-level supervision is important for multi-path reasoning.
\textit{Hallucination might occur with no obstructions.}
In both synthetic and real \textit{No obstructions} settings, Gemini Robotics-ER 1.5 attains low SR (68.7\% and 36.0\%), often hallucinating obstructions even when the target is fully accessible.
Qwen2.5-VL (ICL) performs even worse under the same condition.
\textit{Reasoning quality aligns with final accuracy.}
A clear correlation is observed between MP\_NED and SR-F1, where lower MP\_NED consistently coincides with higher SR across all difficulty levels.
\textit{Limited effectiveness of ICL.}
ICL generally enhances reasoning performance, but it tends to amplify hallucinations in non-obstruction cases.
\textit{Spatial grounding of multi-step reasoning can fail}. 
Baseline models often misalign reasoning steps with object coordinates, causing identity confusion and a high MP\_NED ($>0.8$). 
The low scores (mostly below $10$) in \cref{tab:syn_metric2} and \ref{tab:real_metric2} in the Natural Language Prompting setting indicate poor performance in spatial grounding.

\begin{table*}[t]
\caption{Path-level reasoning results on \ourdataset real set. 
ICL: In-Context Learning.
SFT: Supervised fine tuning.
SR: Success Rate;
MP\_NED: Multi-Path Normalized Edit Distance;
P: Precision;
R: Recall;
F1: F1-Score.
Best result \textbf{bold}; 
second best \underline{underlined}.
}
\vspace{-3mm}
\tabcolsep 2pt
\label{tab:real_metric1}
\resizebox{\textwidth}{!}{
\begin{tabular}{l cc| cccc| cccc |cccc}
\toprule
\multirow{2}{*}{Method} &
\multicolumn{2}{c}{\cellcolor{gray!20}No obstructions} &
\multicolumn{4}{c}{\cellcolor{lightgreen}Easy} &
\multicolumn{4}{c}{\cellcolor{lightyellow}Medium} &
\multicolumn{4}{c}{\cellcolor{verylightpink}Hard} \\
\cmidrule(lr){2-3} \cmidrule(lr){4-7} \cmidrule(lr){8-11} \cmidrule(lr){12-15}
& SR (\%)$\uparrow$ & MP\_NED$\downarrow$
& SR-P (\%)$\uparrow$ & SR-R (\%)$\uparrow$ & SR-F1 (\%)$\uparrow$ & MP\_NED$\downarrow$
& SR-P (\%)$\uparrow$ & SR-R (\%)$\uparrow$ & SR-F1 (\%)$\uparrow$ & MP\_NED$\downarrow$
& SR-P (\%)$\uparrow$ & SR-R (\%)$\uparrow$ & SR-F1 (\%)$\uparrow$ & MP\_NED$\downarrow$ \\
\midrule

\rowcolor{gray!15}
\multicolumn{15}{c}{Oracle (with SoM)} \\

Gemini Robotics-ER 1.5~\cite{abdolmaleki2025gemini}
& 36.0 & 0.85 
& 47.7 & 48.3 & 47.9 & 0.80 
& 40.8 & 28.2 & 32.1 & 0.90 
& 31.8 & \underline{27.3} & 28.8 & 0.91 \\

Gemini Robotics-ER 1.5~\cite{abdolmaleki2025gemini} (ICL)
& 35.0 & 0.83 
& 60.1 & 62.9 & 60.9 & 0.80 
& 44.2 & 34.3 & 36.4 & 0.89 
& \underline{38.6} & \underline{27.3} & \underline{30.6} & 0.92 \\

Qwen2.5-VL~\cite{bai2025qwen2} (ICL)
& 1.3 & 0.56
& 42.4 & 47.2 & 43.5 & 0.37
& 38.3 & 28.2 & 30.0 & 0.68
& 24.4 & 26.1 & 22.7 & \underline{0.79} \\

Qwen2.5-VL~\cite{bai2025qwen2} (SFT)
& \underline{69.7} & -
& \underline{70.2} & \underline{70.6} & \underline{70.4} & -
& \underline{62.4} & \underline{41.6} & \underline{48.0} & -
& \underline{38.6} & 21.6 & 25.9 & - \\
\ourmethod
& \textbf{72.5} & \textbf{0.16}
& \textbf{76.2} & \textbf{79.0} & \textbf{77.2} & \textbf{0.15}
& \textbf{76.6} & \textbf{59.6} & \textbf{64.4} & \textbf{0.40}
& \textbf{79.5} & \textbf{59.1} & \textbf{63.9} & \textbf{0.55} \\

\rowcolor{gray!15}
\multicolumn{15}{c}{Natural Language Prompting} \\

Gemini Robotics-ER 1.5~\cite{abdolmaleki2025gemini}
& 37.7 & 0.85 
& 50.3 & 51.0 & 50.5 & 0.79 
& 47.3 & 32.7 & 37.1 & 0.89 
& \underline{38.9} & \underline{33.3} & \underline{35.2} & \underline{0.89}  \\

Gemini Robotics-ER 1.5~\cite{abdolmaleki2025gemini} (ICL)
& 35.1 & \underline{0.83} 
& 60.5 & 63.3 & 61.4 & 0.80 
& 44.7 & 34.6 & 36.8 & 0.88 
& 38.6 & 27.3 & 30.6 & 0.92  \\

Qwen2.5-VL~\cite{bai2025qwen2} (ICL)
& 10.0 & \underline{0.83}
& 21.3 & 22.6 & 21.7 & 0.79 
& 19.6 & 13.8 & 15.5 & 0.86 
& 11.4 & 13.6 & 12.1 & 0.90 \\

Qwen2.5-VL~\cite{bai2025qwen2} (SFT)
& \textbf{70.0} & - 
& \underline{64.0} & \underline{64.5} & \underline{64.2} & - 
& \underline{61.6} & \underline{37.9} & \underline{45.4} & -
& 29.5 & 19.3 & 21.8 & - \\

\ourmethod
& \textbf{70.0} & \textbf{0.23} 
& \textbf{71.1} & \textbf{71.8} & \textbf{71.3} & \textbf{0.26} 
& \textbf{62.9} & \textbf{40.1} & \textbf{47.3} & \textbf{0.63} 
& \textbf{54.5} & \textbf{35.2} & \textbf{40.3} & \textbf{0.76}  \\

\bottomrule
\end{tabular}
}
\end{table*}

\begin{table*}[t]

\caption{
Object-level reasoning results on \ourdataset real set.
ICL: In-Context Learning;
SFT: Supervised fine tuning;
OP: Object triplet Precision;
OR: Object triplet Recall;
F1\textsubscript{rel}: Object triplet F1-Score;
Best results \textbf{bold}; 
second best \underline{underlined}.
}
\label{tab:real_metric2}
\vspace{-3mm}
\tabcolsep 15pt
\resizebox{\textwidth}{!}{
\begin{tabular}{lccc| ccc| ccc |ccc}
\toprule
\multirow{2}{*}{Method} &
\multicolumn{3}{c}{\cellcolor{lightgreen}Easy} &
\multicolumn{3}{c}{\cellcolor{lightyellow}Medium} &
\multicolumn{3}{c}{\cellcolor{verylightpink}Hard} &
\multicolumn{3}{c}{Overall} \\
\cmidrule(lr){2-4} \cmidrule(lr){5-7} \cmidrule(lr){8-10} \cmidrule(lr){11-13}
& OP$\uparrow$ & OR$\uparrow$ & F1\textsubscript{rel}$\uparrow$
& OP$\uparrow$ & OR$\uparrow$ & F1\textsubscript{rel}$\uparrow$
& OP$\uparrow$ & OR$\uparrow$ & F1\textsubscript{rel}$\uparrow$
& OP$\uparrow$ & OR$\uparrow$ & F1\textsubscript{rel}$\uparrow$ \\
\midrule
\rowcolor{gray!15}
\multicolumn{13}{c}{Oracle (with SoM)} \\

Gemini Robotics-ER 1.5~\cite{abdolmaleki2025gemini}
& 64.9 & 72.9 & 67.1 
& 66.7 & 43.3 & 49.4 
& 41.7 & \underline{19.3} & 24.4 
& 64.9 & 62.9 & 60.8 \\

Gemini Robotics-ER 1.5 (ICL)~\cite{abdolmaleki2025gemini}
& \underline{74.6} & \underline{81.8} & \underline{76.7} 
& \textbf{75.0} & \underline{46.6} & \underline{54.3} 
& \textbf{56.4} & 18.0 & \underline{26.4} 
& \textbf{74.3} & \underline{69.9} & \underline{68.8} \\

Qwen2.5-VL (ICL)~\cite{bai2025qwen2}
& 36.0 & 39.6 & 36.8 
& 36.8 & 18.6 & 23.3 
& 30.9 & 10.2 & 13.1 
& 36.1 & 32.7 & 32.2 \\

\ourmethod
& \textbf{75.7} & \textbf{84.7} & \textbf{78.1} 
& \underline{69.0} & \textbf{57.2} & \textbf{58.2} 
& \underline{55.6} & \textbf{43.9} & \textbf{44.7}
& \underline{73.2} & \textbf{75.5} & \textbf{71.4} \\

\rowcolor{gray!15}
\multicolumn{13}{c}{Natural Language Prompting} \\

Gemini Robotics-ER 1.5~\cite{abdolmaleki2025gemini}
& 3.7 & 4.0 & 3.8 
& 2.9 & 1.4 & 1.8 
& 10.2 & 2.6 & 3.8 
& 3.6 & 3.2 & 3.2 \\

Gemini Robotics-ER 1.5~\cite{abdolmaleki2025gemini} (ICL)
& 3.2 & 3.6 & 3.4 
& 4.8 & 2.5 & 3.2 
& 9.1 & 1.7 & 2.8 
& 3.8 & 3.2 & 3.3 \\

Qwen2.5-VL~\cite{bai2025qwen2} (ICL)
& \underline{5.9} & \underline{6.9} & \underline{6.2} 
& \underline{5.1} & \underline{2.7} & \underline{3.3} 
& \underline{11.4} & \underline{3.2} & \underline{4.9} 
& \underline{5.8} & \underline{5.6} & \underline{5.3} \\

\ourmethod
& \textbf{56.9} & \textbf{58.9} & \textbf{57.5} 
& \textbf{47.0} & \textbf{26.1} & \textbf{32.1} 
& \textbf{35.6} & \textbf{15.6} & \textbf{20.6} 
& \textbf{53.5} & \textbf{48.2} & \textbf{49.1} \\

\bottomrule
\end{tabular}
}
\end{table*}

\subsection{Ablation studies}\label{sec:ablations}

\begin{table}[t]
\caption{Ablation study on SFT on synthetic set (Overall).}
\label{tab:sft-occlusion-ablation-overall}
\vspace{-3mm}
\tabcolsep 15pt
\resizebox{\columnwidth}{!}{%
\begin{tabular}{lccc}
\toprule
Method & SR-F1 $\uparrow$ & OR-F1 $\uparrow$ & MP\_NED $\downarrow$ \\
\midrule
Baseline
& 74.7 & 71.9 & 0.220 \\

+ Contact point 
& 75.3 \textcolor{brightgreen}{(+0.6)}
& 72.5 \textcolor{brightgreen}{{(+0.6)}}
& 0.216 \textcolor{brightgreen}{{(-0.004)}} \\

+ Degree word 
& 75.1 \textcolor{brightgreen}{{(+0.4)}}
& 72.5 \textcolor{brightgreen}{{(+0.6)}}
& 0.217 \textcolor{brightgreen}{{(-0.003)}} \\

+ Occlusion ratio 
& \textbf{76.4} \textcolor{brightgreen}{{(+1.7)}}
& \textbf{73.3} \textcolor{brightgreen}{{(+1.4)}}
& \textbf{0.210} \textcolor{brightgreen}{{(-0.010)}} \\

\bottomrule
\end{tabular}
}
\end{table}

\noindent\textbf{SFT with obstruction information.}
\cref{tab:sft-occlusion-ablation-overall} shows results when adding obstruction cues during SFT instead of using only the obstruction graph.
Contact point, Degree word, and Occlusion ratio cues improve success rates across all difficulty levels, with the largest gains on Hard cases.
Ratio yields the most significant improvement ($+5.8\%$ SR-F1 on Hard), increasing precision and decreasing the reasoning error. The full table is provided in \textit{Supp. Mat.}

\begin{table*}[t]
\caption{
Ablation study on RFT (Synthetic set). The harder the scene is, the higher the contribution of RFT.}
\vspace{-3mm}
\label{tab:rl-reward-ablation}
\tabcolsep 6pt
\resizebox{\textwidth}{!}{%
\begin{tabular}{l ccc|ccc|ccc|ccc}
\toprule
\multirow{2}{*}{Variant} &
\multicolumn{3}{c}{\cellcolor{lightgreen}Easy} &
\multicolumn{3}{c}{\cellcolor{lightyellow}Medium} &
\multicolumn{3}{c}{\cellcolor{verylightpink}Hard} &
\multicolumn{3}{c}{Overall} \\
\cmidrule(lr){2-4} \cmidrule(lr){5-7} \cmidrule(lr){8-10} \cmidrule(lr){11-13}
& SR-F1$\uparrow$ & OR-F1$\uparrow$ & MP\_NED$\downarrow$
& SR-F1$\uparrow$ & OR-F1$\uparrow$ & MP\_NED$\downarrow$
& SR-F1$\uparrow$ & OR-F1$\uparrow$ & MP\_NED$\downarrow$
& SR-F1$\uparrow$ & OR-F1$\uparrow$ & MP\_NED$\downarrow$ \\
\midrule

Baseline (SFT)
& 81.8 & 80.9 & 0.115
& 67.1 & 59.4 & 0.389
& 50.1 & 46.9 & 0.525
& 76.4 & 73.3 & 0.210 \\

+ RFT on Answer
& 83.3\;\textcolor{brightgreen}{{(+1.5)}} & 82.6\;\textcolor{brightgreen}{{(+1.7)}} & 0.109\;\textcolor{brightgreen}{{(-0.006)}}
& 69.1\;\textcolor{brightgreen}{{(+2.0)}} & 62.0\;\textcolor{brightgreen}{{(+2.6)}} & 0.370\;\textcolor{brightgreen}{{(-0.019)}}
& 54.5\;\textcolor{brightgreen}{{(+4.4)}} & 48.7\;\textcolor{brightgreen}{{(+1.8)}} & 0.507\;\textcolor{brightgreen}{{(-0.018)}}
& 78.2\;\textcolor{brightgreen}{{(+1.8)}} & 75.3\;\textcolor{brightgreen}{{(+2.0)}} & 0.201\;\textcolor{brightgreen}{{(-0.009)}} \\

\bottomrule
\end{tabular}
}
\end{table*}

\noindent\textbf{RFT with obstruction-aware reward.}
\cref{tab:rl-reward-ablation} shows how results change when apply a set-level IoU reward to the predicted answers during RFT. 
This yields to consistent improvements across all metrics, with SR-F1 gains increasing with complexity: Easy ($+1.5\%$), Medium ($+2.0\%$), and Hard ($+4.4\%$). 
Greater improvements under severe obstruction indicate the IoU reward effectively encourages complete answer sets, proving beneficial over binary correctness when multiple ground-truth obstructors exist. 
Though applied only to \texttt{<answer>} output, reasoning metrics (OR-F1 and MP\_NED) also improve, suggesting that promoting complete answers indirectly guides more faithful reasoning traces.

\subsection{Qualitative analysis}\label{sec:qualitative_analysis}

\begin{figure*}[t]
\begin{center}
  \begin{tabular}{@{\,}c@{\,}c@{\,}c@{\,}c@{\,}c}
    \begin{overpic}[width=.197\textwidth]{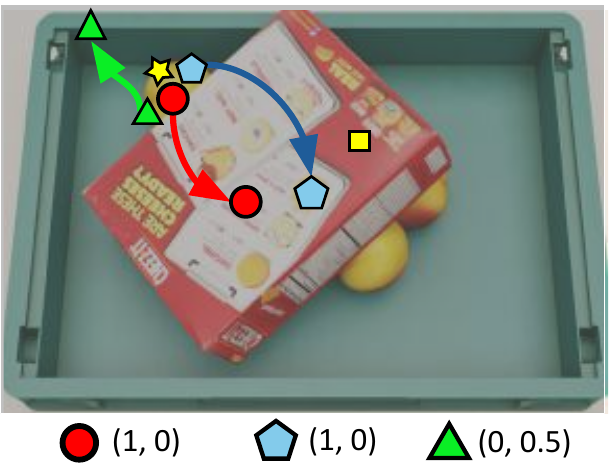}
      \put(43,78){\color{black}\scriptsize\textbf{Easy}}
      \put(-8,20){\rotatebox{90}{\color{black}\scriptsize\textbf{Synthetic}}}
    \end{overpic}&
    \begin{overpic}[width=.197\textwidth]{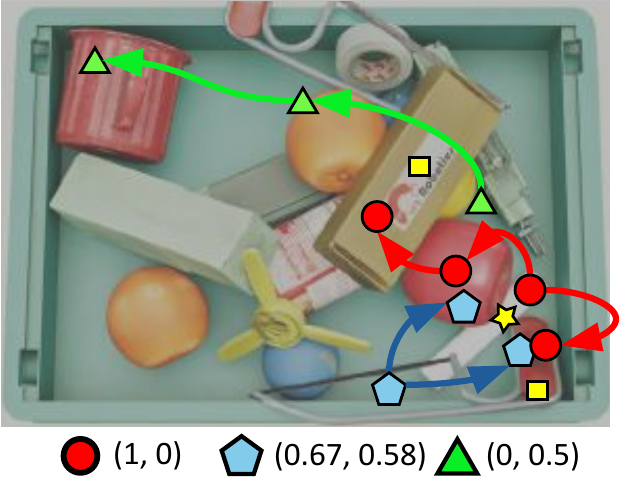}
      \put(35,78){\color{black}\scriptsize\textbf{Medium}}
    \end{overpic}&
    \begin{overpic}[width=.188\textwidth]{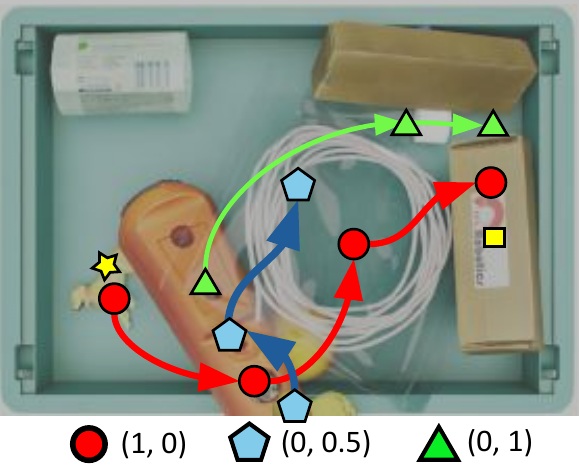}
      \put(38,82){\color{black}\scriptsize\textbf{Hard}}
    \end{overpic}&
    \begin{overpic}[width=.188\textwidth]{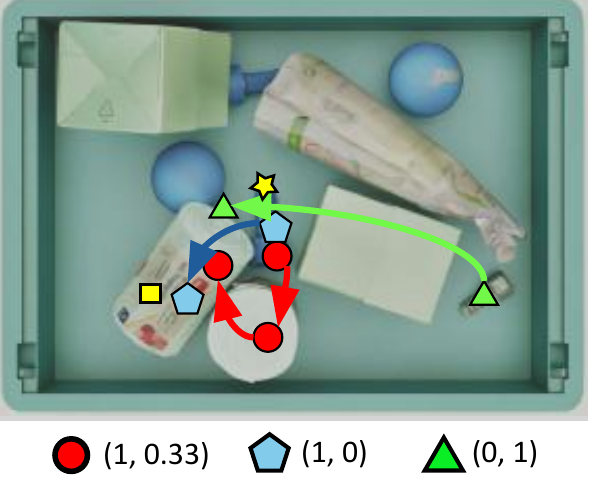}
      \put(25,82){\color{black}\scriptsize\textbf{Reasoning Err.}}
    \end{overpic}&
    \begin{overpic}[width=.193\textwidth]{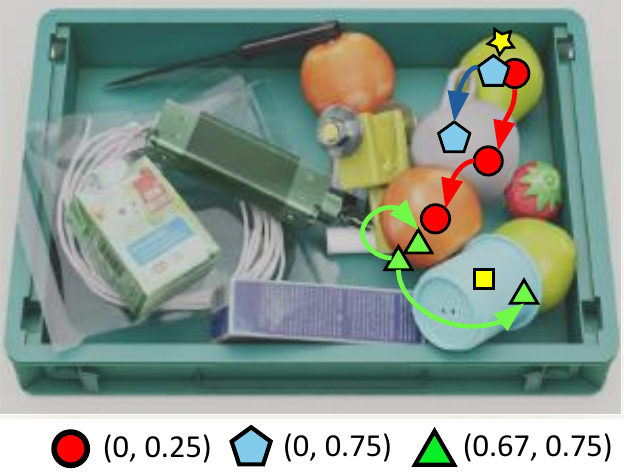}
      \put(12,80){\color{black}\scriptsize\textbf{Reasoning + Ans. Err.}}
    \end{overpic}\\
    \begin{overpic}[width=.19\textwidth]{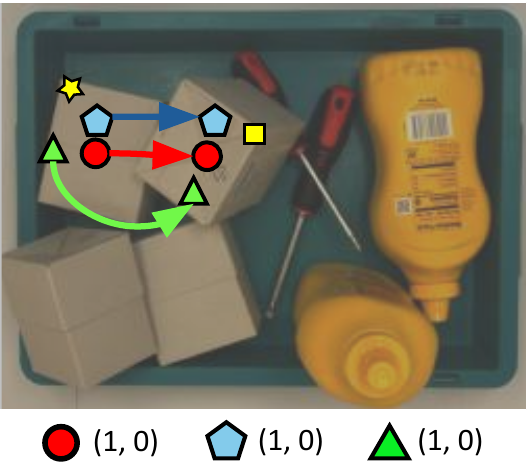}
        \put(-8,30){\rotatebox{90}{\color{black}\scriptsize\textbf{Real}}}
    \end{overpic}&
    \begin{overpic}[width=.19\textwidth]{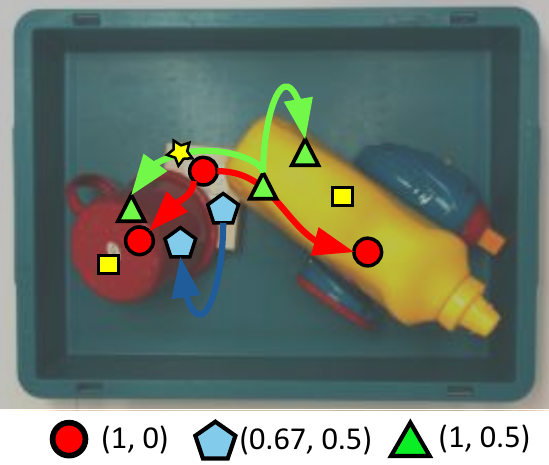}
    \end{overpic}&
    \begin{overpic}[width=.19\textwidth]{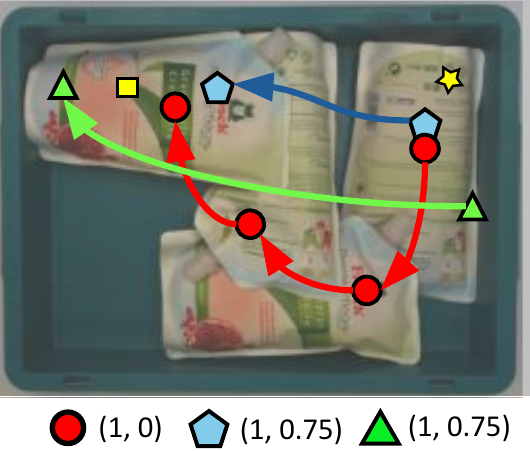}
    \end{overpic}&
    \begin{overpic}[width=.19\textwidth]{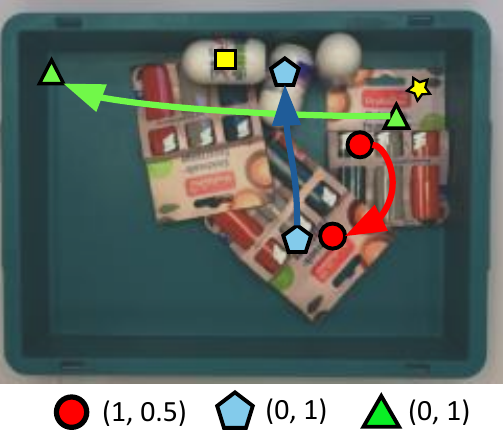}
    \end{overpic}&
    \begin{overpic}[width=.19\textwidth]{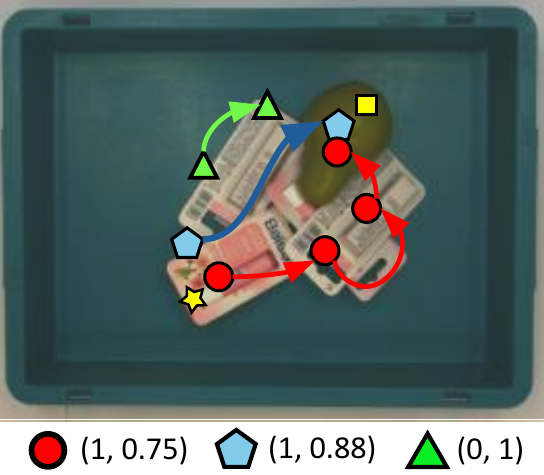}
    \end{overpic}\\
    \\
  \end{tabular}
\end{center}
\vspace{-10mm}
\caption{Qualitative results on \ourdataset different splits, and in two types of failure.
\includegraphics[height=0.9em]{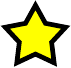} mark the target object, \includegraphics[height=0.9em]{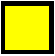} the top obstructor, \includegraphics[height=0.9em]{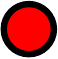}~UNOGrasp, \includegraphics[height=0.9em]{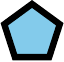}~Gemini Robotics-ER~1.5, 
and \includegraphics[height=0.9em]{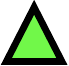}~Qwen2.5-VL (ICL) predictions with their reasoning traces.
$(\text{SR-F1} / \text{MP\_NED})$ scores are reported at the bottom of each image.}
\label{fig:qualitatives}
\end{figure*}

\cref{fig:qualitatives} visualizes obstruction reasoning traces produced by \ourmethod, Gemini Robotics-ER1.5 (ICL)\cite{abdolmaleki2025gemini}, and Qwen2.5-VL (ICL)\cite{bai2025qwen2} in both synthetic and real settings. For each setting, we present examples across the three difficulty levels, along with two failure modes: incorrect reasoning but correct answers, and failures in both.
We observe that Qwen2.5-VL (ICL) struggles with spatial grounding even on easy cases. Gemini tends to terminate prematurely during the multi-step reasoning in complex scenarios, missing top-level obstructors when multiple obstruction paths exist. In contrast, \ourmethod handles multi-path reasoning effectively but can fail with visually similar objects or densely clustered groups.

\begin{figure}[t!]
\begin{center}
  \begin{tabular}{@{}c@{\,}c}
    \begin{overpic}[width=.48\columnwidth]{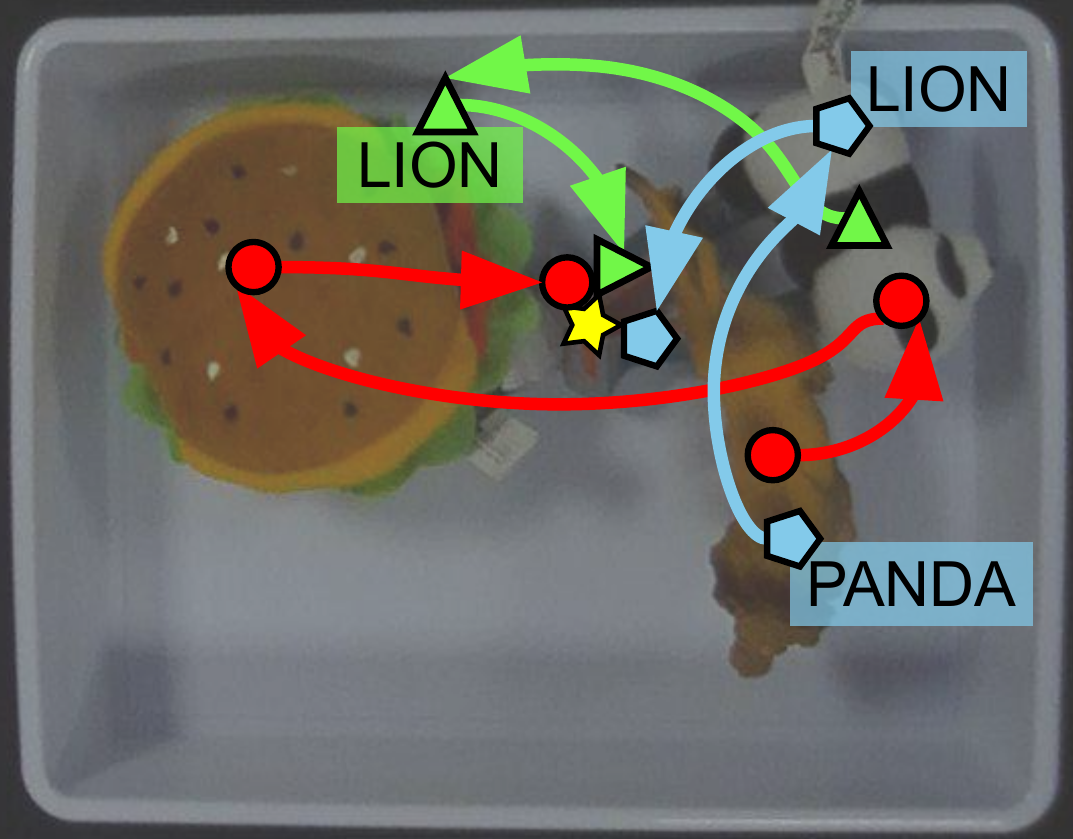}
      \put(25,80){\color{black}\scriptsize\textbf{Medium - Pliers}}
    \end{overpic}&
    \begin{overpic}[width=.48\columnwidth]{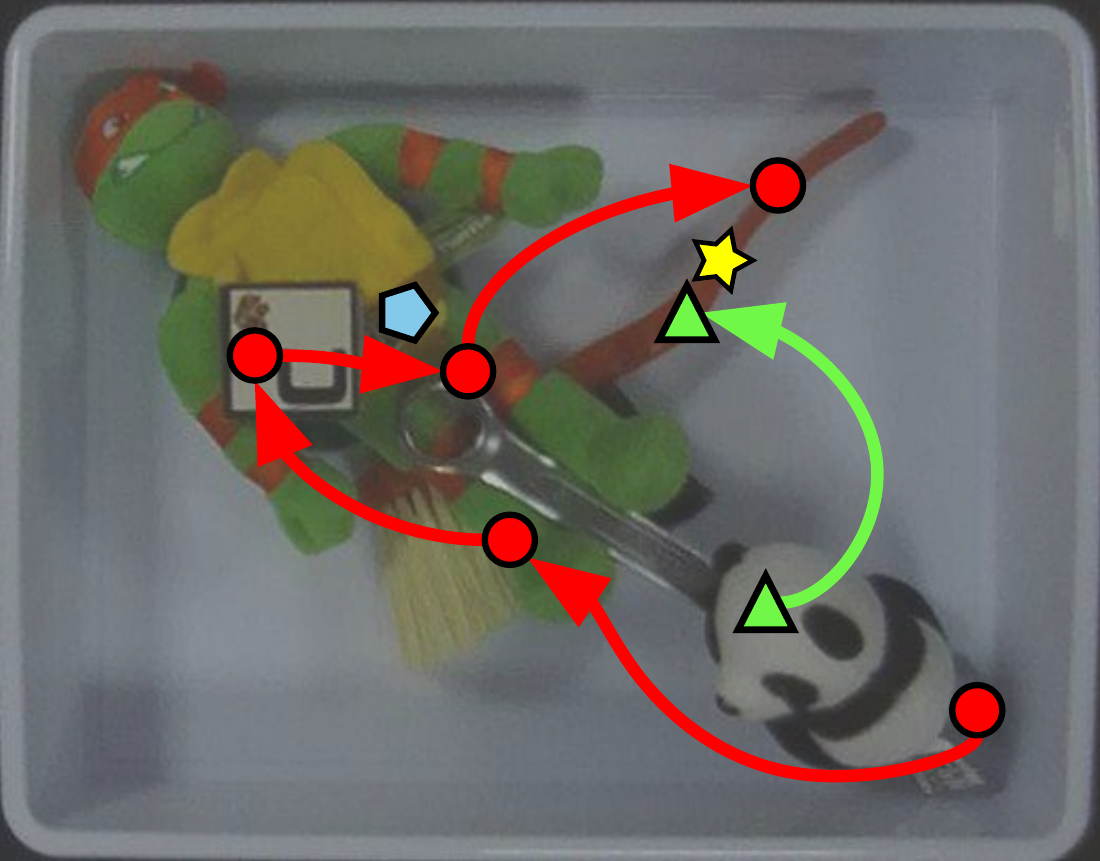}
      \put(26,80){\color{black}\scriptsize\textbf{Hard - Red brush}}
    \end{overpic}\\
  \end{tabular}
\end{center}
\vspace{-7mm}
\caption{Qualitative results from laboratory robotics experiments. 
\includegraphics[height=0.9em]{main/figures/icon/star_compressed.pdf} mark the target object, \includegraphics[height=0.9em]{main/figures/icon/red_compressed.pdf}~UNOGrasp, \includegraphics[height=0.9em]{main/figures/icon/blue_compressed.pdf}~Gemini Robotics-ER~1.5, 
and \includegraphics[height=0.9em]{main/figures/icon/green_compressed.pdf}~Qwen2.5-VL (ICL).
Labels are shown for misaligned predictions (labels-spatial location disagreement). 
Difficulty level and target prompt are display at the top of the figure.}
\label{fig:robot}
\vspace{-5mm}
\end{figure}

\subsection{Laboratory robotic experiments}

We validate UNOGrasp on a UR5e robotic platform across 30 real-world scenarios with 25 distinct objects. 
Objects are placed in a bin and captured by a top-down ZED 2 camera. 
We compare against Qwen2.5-VL and Gemini Robotics-ER 1.5 (both ICL), using GroundedSAM~\cite{ren2024groundedSAM} and GraspNet~\cite{Gilles2024MetaGraspNetV2} for grasp pose prediction~\cite{jiao2025free}. 
Methods predict and execute the next object removal.
We measure the success ratio, deeming the action successful when all the objects are removed in the correct order to achieve the final grasping goal.

\noindent In Tab.~\ref{tab:robotic_experiment}, UNOGrasp matches Gemini Robotics-ER 1.5 performance on Easy and Medium, despite training on open-source synthetic data, and significantly outperforms it on Hard (+30\%).
Real-world conditions can feature high-contrast situations (Fig.~\ref{fig:robot}), where black backgrounds and white bins create over-exposure conditions to which Gemini appears more robust. 
Both baselines exhibit similar failure patterns identified in \S\ref{sec:qualitative_analysis}, \ie, regarding spatial grounding, which become more frequent in longer obstruction chains. 
Qwen2.5-VL demonstrates a tendency to over-reason, predicting the container itself as an obstruction even in trivial target-only scenarios.

\begin{table}[t]
\caption{
Real-world robotic experiment with the UR5e, success ratios across 30 scenarios with 25 objects.
}
\label{tab:robotic_experiment}
\tabcolsep 8pt
\vspace{-2mm}
\resizebox{\columnwidth}{!}{%
\begin{tabular}{lcccc}
\toprule
Method & \cellcolor{lightgreen}Easy & \cellcolor{lightyellow}Medium & \cellcolor{verylightpink}Hard & Average \\
\midrule
Gemini Robotics-ER 1.5 \cite{abdolmaleki2025gemini} & \textbf{80\%} & \textbf{30\%} & 10\%          & 40\% \\
Qwen2.5-VL \cite{bai2025qwen2} & 10\%          & 0\%           & 0\%           & 3\% \\
UNOGrasp & \textbf{80\%} & \textbf{30\%} & \textbf{40\%} & \textbf{50\%} \\
\bottomrule
\end{tabular}
}
\vspace{-3mm}
\end{table}

\section{Conclusions}\label{sec:conclusion}

We addressed the critical limitation of VLMs in obstruction reasoning, which can be used for robotic grasping.
Our contributions are twofold: we introduced \ourdataset, a novel dataset, based on MetaGraspMetV2 \cite{Gilles2024MetaGraspNetV2}, featuring $100k+$ annotated obstruction paths for developing and benchmarking; and we proposed \ourmethod, a VLM trained via sequential SFT and RFT with novel visually-grounded and obstruction-aware rewards. 
Experiments showed \ourmethod significantly improves obstruction reasoning, achieving 78.2\% precision on average on \ourdataset, and 50\% success rate on laboratory robotics experiments with a setup (\eg, objects, camera) different from that of finetuning data.
Importantly, real-robot evaluations confirm the advantage of \ourmethod over existing generalist and proprietary model in difficult scenes, with good generalization capability despite being trained only with synthetic data.
Future work can focus on scaling \ourmethod and \ourdataset to incorporate multi-view perception and further expand the object diversity.

\subsection*{Acknowledgements} This work was supported by PNRR ICSC National Research Centre for HPC, Big Data and Quantum Computing (CN00000013), and FAIR - Future AI Research (PE00000013), funded by NextGeneration EU.
We also acknowledge ISCRA for granting us access to the LEONARDO supercomputer, owned by the EuroHPC Joint Undertaking and hosted by CINECA (Italy).

{
    \small
    \bibliographystyle{main/ieeenat_fullname}
    \bibliography{main}
}
\clearpage
\setcounter{page}{1}
\maketitlesupplementary
\renewcommand{\thesection}{\Alph{section}}

\setcounter{section}{0}
\setcounter{figure}{0}
\setcounter{table}{0}

\renewcommand{\thetable}{\Alph{table}}
\renewcommand{\thefigure}{\Alph{figure}}

In this supplementary material, we provide more details regarding our benchmark~\ourdataset (\cref{sec:add_bench}) and the training details of our method~\ourmethod (\cref{sec:add_method}). We also present more additional experimental details and analyses on our benchmark (\cref{sec:add_imp}) and the real-robot evaluation (\cref{sec:add_robot}). Finally we show a video (unograspRoboticResults.mp4) demonstrating the real-robot test with Qwen2.5-VL~\cite{bai2025qwen2}, Gemini Robotics-ER 1.5~\cite{abdolmaleki2025gemini} and \ourmethod under \inlineColorbox{lightgreen}{Easy}, \inlineColorbox{lightyellow}{Medium} and \inlineColorbox{lightred}{Hard} scenarios.

\section{Additional details on \ourdataset}
\label{sec:add_bench}

\subsection{Dataset construction and statistics}

We construct the Synthetic subset of \ourdataset starting from 8{,}007 scenes in MetaGraspNetV2 (37 viewpoints each) and apply a series of pre-processing steps.
We first filter out scenes that are inappropriate for our benchmarking purpose: i) we remove empty scenes or scenes containing only a single object, \ie no obstruction can be formed; ii) we remove scenes with obstructions that are not physically plausible (\eg, objects penetrating each other due to simulation artifacts); iii) we discard scenes that contain bidirectional or cyclic obstruction patterns, as they will inevitably cause problematic obstruction paths.

After the filtering, we retain 6{,}255 scenes, from which we randomly sample four diverse viewpoints per scene, resulting in a final set of 25{,}020 images.
Almost all objects serve as target objects for constructing the obstruction graphs and generating the VQA dataset. We exclude some objects when i) their obstruction ratios are below 1\% as such light obstruction are often visually ambiguous and do not impede grasping execution, and ii) when their obstruction ratios above 95\% as they become visually difficult to recognize due to such excessive obstruction.
For the Real subset, we follow the same construction pipeline as in the Synthetic subset. As mentioned in \cref{sec:exps}, we split the synthetic scenes into training, validation, and testing sets with a 7:1:2 ratio, while all the real scenes are exclusively used for testing. 
\cref{tab:data_count} summarizes the dataset statistics of all subsets in UNOBench. Note that for the synthetic subset, some objects are annotated by human as ``indescribable objects'' as they are barely visible due to severe obstruction.
We thus remove all VQA samples containing such objects in the NLP setting, resulting in fewer objects compared to the Oracle (SoM) setting. 
In addition, we randomly sample 2{,}000 out of 12{,}719 objects under Oracle (SoM) Test that have No Obstruction (No-Obs) for evaluation efficiency.

\begin{table}[h!]
\centering
\caption{Statistics of the Oracle (SoM) and Natural Language Prompting (NLP) settings on both Synthetic and Real subsets.}
\label{tab:data_count}
\vspace{-4pt}
\small
\resizebox{\columnwidth}{!}{%
\begin{tabular}{lcccccc}
\toprule
\textbf{Setting} & \textbf{Split} & \textbf{\#Objects} & \textbf{No-Obs} & \textbf{Easy} & \textbf{Med} & \textbf{Hard} \\
\midrule

\rowcolor{gray!15}
\multicolumn{7}{c}{\textbf{Synthetic}} \\
\midrule

\multirow{3}{*}{\textbf{Oracle (SoM)}} 
& Train & 67{,}945 & 43{,}350 & 16{,}553 & 7{,}305 & 737 \\
& Val   & 9{,}539  & 6{,}242  & 2{,}260  & 974   & 63 \\
& Test  & 8{,}863  & 2{,}000  & 4{,}576  & 2{,}085 & 202 \\
\cmidrule(lr){1-7}

\multirow{3}{*}{\textbf{NLP}} 
& Train & 61{,}690 & 40{,}389 & 14{,}801 & 5{,}969 & 531 \\
& Val   & 8{,}785  & 5{,}853  & 2{,}056  & 830   & 46 \\
& Test  & 8{,}526  & 1{,}993  & 4{,}477  & 1{,}936 & 180 \\

\midrule

\rowcolor{gray!15}
\multicolumn{7}{c}{\textbf{Real (test only)}} \\
\midrule

\textbf{Oracle (SoM)} 
& Test & 2{,}232 & 1{,}341 & 606 & 263 & 22 \\
\textbf{NLP} 
& Test & 2{,}232 & 1{,}341 & 606 & 263 & 22 \\

\bottomrule
\end{tabular}
}
\end{table}

\subsection{Natural-language annotation pipeline}
\begin{figure*}[t]
    \centering

    \begin{minipage}[t]{0.58\textwidth}
        \vspace{8pt}
        \centering
        {\small \textbf{(a) Input image}}
        \vspace{15pt}
        \includegraphics[width=\linewidth]{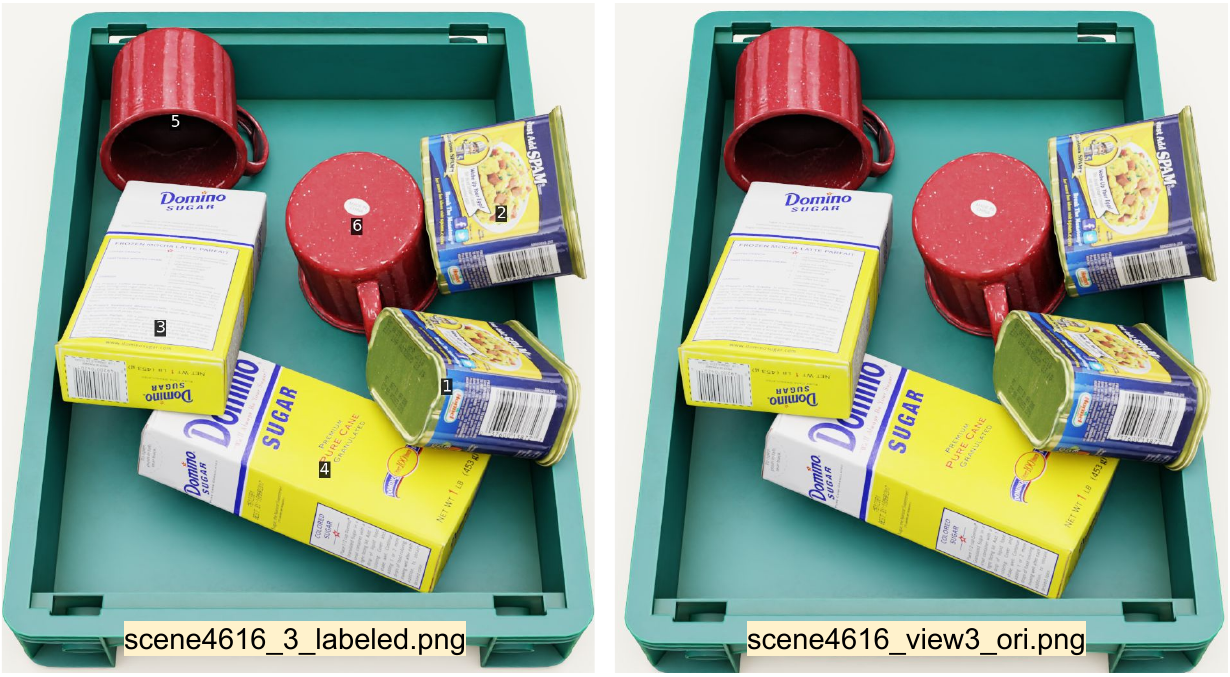}
    \end{minipage}
    \hfill
    \begin{minipage}[t]{0.38\textwidth}
        {\small \textbf{(b) Original obstruction graph (OG)}}
        \vspace{4pt}
\begin{lstlisting}[basicstyle=\ttfamily\scriptsize]
{
  "scene_id": "data_ifl_46/scene4616",
  "view_id": "3",
  "target_object": 4,
  "obstruction_paths": [[1,4], [3,4]],
  "top_objects": [1,3],
  "depends_on": [1,3],
  "k_min": 1,
  "num_paths": 2,
  "new_difficulty": "Medium"
}

{
    "obj1": 3,
    "obj2": 4,
    "relation": "3 occludes 4",
    "mask_ratio": 0.0681,
    "point": {"x": 454, "y": 686}
}

\end{lstlisting}

    \end{minipage}

\begin{minipage}[t]{\textwidth}
    {\small \textbf{(d) VQA Dataset (Oracle with SoM)}}
    \vspace{-2pt}

\begin{lstlisting}[basicstyle=\ttfamily\scriptsize]
(*\textbf{Image:}*) "Meta_reason_data/scene4616_3_labeled.png"

(*\textbf{System prompt}:*)
You are an assistant for robotic grasp planning. When asked which object must be removed first to grasp a specific object:
- If the target object is not obstructed, return the target object's ID itself.
- If there is one obstruction path, reason step-by-step along that path, include occlusion ratios when available, and end with the top-most object.
- If multiple obstruction paths exist, reason step-by-step for each path separately.
- In the <answer>...</answer> tag, output ALL distinct top-most objects as a JSON list.
Use <think>...</think> for reasoning, and put ONLY final object IDs inside <answer>...</answer>.

(*\textbf{Human question}:*)
"<image>\nTo grasp object 4, which object is on top of it?"

(*\textbf{Answer}:*)
"<think>
Path1: Object 4 is obstructed by object 1 with the occlusion ratio of 4%
Path2: Object 4 is obstructed by object 3 with the occlusion ratio of 7%
</think>
<answer>[1, 3]</answer>"
\end{lstlisting}

\end{minipage}

\vspace{0pt}

{\scriptsize
\begin{minipage}[t]{\textwidth}
    {\small \textbf{(e) VQA Dataset (Natural Language Prompting)}}
    \vspace{-2pt}
\begin{lstlisting}[basicstyle=\ttfamily\scriptsize, escapeinside={(*}{*)}]
(*\textbf{Image:}*)
"Meta_reason_data_ori/scene4616_view3_ori.png"

(*\textbf{System prompt}:*)
You are an assistant specialized in robotic grasp planning based on obstruction reasoning.
When asked which object must be removed first to grasp a specific target object in a single image:
- If the target object is not obstructed, return the target object's name/description and its coordinates.
- If the object has one obstruction path, reason step-by-step along that path, include the occlusion ratio for each obstruction relation when available, and end with the top-most object. Each reasoning step must reference explicit (x,y) coordinates.
- If the object has multiple obstruction paths, reason step-by-step for each path separately, and include ALL distinct top-most occluding objects in the final answer.
- All reasoning must be enclosed inside a single pair of <think>...</think>.
- The final answer must be enclosed in <answer>...</answer> and must follow the format:
  <answer>[<points x y>object name</points>, ...]</answer>

(*\textbf{Human question}:*)
"<image>\nTo grasp right sugar box, which object is on top of it?"

(*\textbf{Answer}:*)
"<think>Path1: right sugar box at (640, 807) is obstructed by bottom canned meat at (831, 677) with the occlusion ratio of 4%
<answer>[<points 831 677>bottom canned meat</points>, <points 383 585>left sugar box</points>]</answer>"
\end{lstlisting}

\end{minipage}
}
\vspace{-10pt}

    \caption{
    \textbf{UNOBench example.}
    Illustration of scene representation and corresponding VQA formulations.
    }
    \label{fig:dataset_example_full}
\end{figure*}

Fig.~\ref{fig:dataset_example_full} presents a complete example from UNOBench, including the input image, the corresponding obstruction graph, pairwise obstruction relations, and the generated VQA samples in both the Oracle (SoM) and NLP settings. 
Below, we present the detailed procedure for obtaining the high-quality object names in natural language for the NLP setting. 

\noindent\textbf{LLM pre-annotation.}
We first perform an automated pre-annotation with a vision-language model (VLM), \ie, gpt-4o. Specifically, given the SoM-labeled images, the VLM first produces one short natural-language description per object. We instruct the VLM to consider, especially in scenes containing multiple instances of the same semantic category, to produce names using spatial specifiers (\eg, left, right, top, bottom) to uniquely distinguish multiple instances of the same category. Moreover, we instruct the output to follow a shared JSON format to ease the parsing at later stages.

\noindent\textbf{Prolific human annotation.}
For scenes containing more than five objects, we further conduct human annotation on all objects, where the VLM-generated descriptions are revised and corrected by human annotators.
To collect human annotations, we used a commercial platform, Prolific\footnote{https://www.prolific.com/}, for its high-quality annotation and ethical compliance. Specifically, we recruited annotators who are native or primary English speakers located in six English-speaking countries, with a $\geq 99\%$ historical approval rate on the platform, under the recommended pay rate.
A total of 196 annotators participated. Each annotator could complete at most 45 scenes, and the study was designed to be finished within 80 minutes to prevent fatigue. The participant pool had a balanced gender distribution and was mainly from the UK (54\%), the US (27\%), and Canada (12\%), with birth countries spanning 11 regions.

\begin{figure*}[t]
    \centering
    \includegraphics[width=0.95\textwidth]{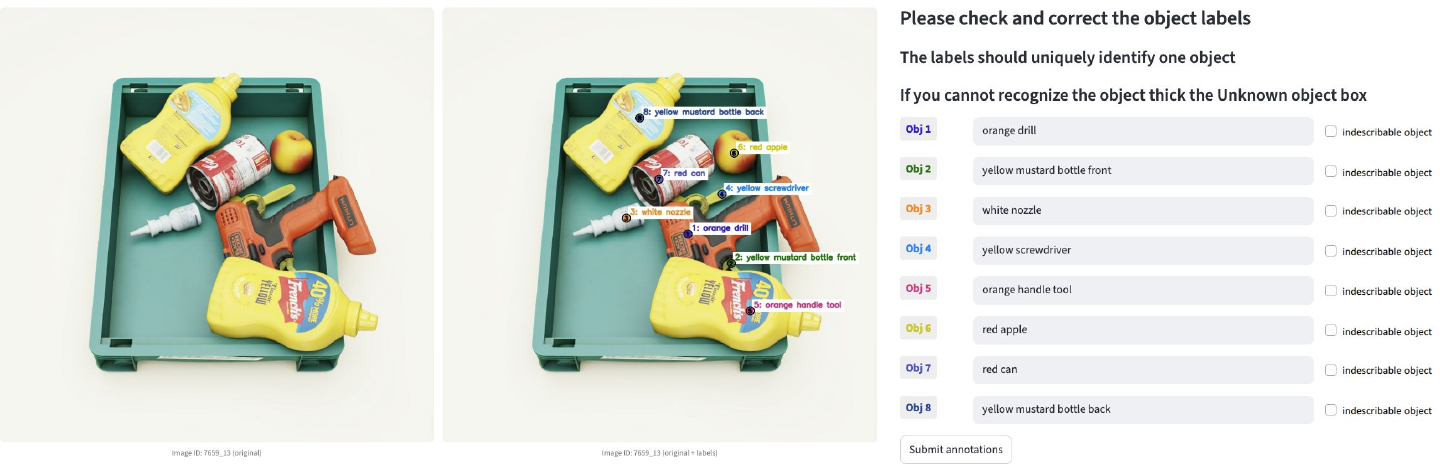}
    \caption{Interface of the online annotation platform.}
    \label{fig:anno_app}
\end{figure*}

Our annotations were collected via a web interface where we present to the annotators the scene with colored dots, each dot marks an object and its VLM-generated preliminary name in the same color next to the dot (as shown in Fig.~\ref{fig:anno_app}).
Annotators are requested to verify whether each provided name uniquely and accurately referred to the marked object. In case no, they should correct the name accordingly. Under cases where objects are barely visible due to heavy obstruction, annotators can mark them as ``indescribable object''.

Moreover, we further conducted a review on top of the collected annotations from Prolific, to further check for semantic or referential inconsistencies. 
In total, 5{,}400 images and 41{,}193 object names were examined; 4{,}78 images and 17{,}261 names were corrected, and the remaining names were confirmed to be accurate.

\subsection{Evaluation metrics}

As introduced in \cref{sec:eval_procedure}, we evaluate model predictions at three levels, including i) the outcome-level metrics in terms of how accurate we can identify top obstructors, ii) object-level reasoning in terms of how accurate we can predict pairwise obstruction relationships, and iii) path-level reasoning in terms of how accurate all obstruction paths are predicted. Algorithm~\ref{alg:evaluation} outlines the evaluation algorithm. For each sample, we extract the predicted top obstructors $\mathcal{F}_{\text{pred}}$ and predicted reasoning paths $\mathcal{A}_{\text{pred}}$ from \texttt{<answer>} and \texttt{<think>} respectively, and load the corresponding ground-truth sets $\mathcal{F}_{\text{gt}}$ and $\mathcal{A}_{\text{gt}}$.

\noindent\textbf{Outcome-level metrics.}
Since a target may have multiple top-level obstructors, we compute precision, recall, and F1 between $\mathcal{F}_{\text{pred}}$ and $\mathcal{F}_{\text{gt}}$ to evaluate whether the model correctly identifies all required top-level blockers without missing or introducing wrong objects.

\noindent\textbf{Object-level reasoning.}
We extract all pairwise obstruction triplets from predicted and ground truth paths to form $T_{\text{pred}}$ and $T_{\text{gt}}$. 
OP/OR/F1\textsubscript{rel} measure the accuracy of these fundamental obstruction edges, independent of path depth or multi-path structure. 
As shown in Tabs.~\ref{tab:syn_metric2} and \ref{tab:real_metric2}, existing VLMs still struggle with this basic reasoning ability.

\noindent\textbf{Path-level reasoning.}
Let $\mathcal{P}=\{p_1,\dots,p_m\}$ and $\mathcal{G}=\{g_1,\dots,g_n\}$ denote the predicted and ground-truth path sets, respectively. 
For each pair $(p_i,g_j)$, we compute the normalized graph edit distance:
\[
\text{NED}(p_i,g_j) \;=\; 
\frac{\text{Levenshtein}(p_i,g_j)}{\max(|p_i|,|g_j|)} 
\in [0,1].
\]
This distance captures three fundamental reasoning errors (Fig.~\ref{fig:dataset}): substitutions (incorrect relations), deletions (missing objects), and insertions (extra objects). 
If $m<n$ (under prediction), we append $(n-m)$ dummy predictions with a fixed missing penalty (weight set to 1); if $m>n$ (over-prediction), we append $(m-n)$ dummy ground-truth paths with a redundant penalty (also 1). 
We then apply the Hungarian algorithm to find the minimum-cost matching between $P$ and $G$, and report the average matching cost as MP\_NED, which measures full multi-path consistency.

\begin{algorithm}[t]
\caption{Evaluation Algorithm}
\label{alg:evaluation}
\begin{algorithmic}
\small

\State \textbf{Require:} Predictions $\mathcal{P}$, Ground-truth $\mathcal{G}$
\State \textbf{Ensure:} SR-P/R/F1, OP/OR/F1, MP\_NED

\Statex 1:~\textbf{Parse and Load Data}
\For{each sample $(p, g)$ in $(\mathcal{P}, \mathcal{G})$}
    \State $\mathcal{F}_{\text{pred}}, \mathcal{A}_{\text{pred}} \gets$ Extract sets from $p.\texttt{<answer>}$ and $p.\texttt{<think>}$
    \State $\mathcal{F}_{\text{gt}}, \mathcal{A}_{\text{gt}} \gets$ Load $g[\texttt{top\_objects}]$ and $g[\texttt{obs\_paths}]$

    \Statex 2:~\textbf{Outcome-level Metrics}
    \State $(SR\_P, SR\_R, SR\_F1) \gets \text{ComputePRF}(\mathcal{F}_{\text{pred}}, \mathcal{F}_{\text{gt}})$

    \Statex 3:~\textbf{Reasoning-level Metrics}
    \State \textbf{Object-level Reasoning:}
    \State $T_{\text{pred}} \gets$ Extract triplets $(obj_a, rel, obj_b)$ from $\mathcal{A}_{\text{pred}}$
    \State $T_{\text{gt}} \gets$ Extract triplets from $\mathcal{A}_{\text{gt}}$
    \State $(OP, OR, F1_{\text{rel}}) \gets \text{ComputePRF}(T_{\text{pred}}, T_{\text{gt}})$

    \State \textbf{Path-level Reasoning (MP\_NED):}
    \State $m, n \gets |\mathcal{A}_{\text{pred}}|, |\mathcal{A}_{\text{gt}}|$
    \State Initialize cost matrix $C \in \mathbb{R}^{m \times n}$
    \For{$i \gets 1 \text{ to } m, \ j \gets 1 \text{ to } n$}
        \State $C_{ij} \gets \text{Levenshtein}(p_i, g_j) \ / \ \max(|p_i|, |g_j|)$
    \EndFor
    
    \State $\text{matches} \gets \text{HungarianMatch}(C)$
    \State $\text{MP\_NED} \gets \frac{1}{\max(m, n)} \sum_{(i,j) \in \text{matches}} C_{ij}$

    \State Store all computed metrics
\EndFor

\State \textbf{return} Mean metrics over all samples

\end{algorithmic}
\end{algorithm}

\section{Additional details on \ourmethod}
\label{sec:add_method}

\subsection{SFT details}
\label{sec:add_sft}

To warm-start the model's obstruction reasoning capability, we first optimize $f_{\Theta}$ with supervised fine-tuning (SFT) on the synthetic set of \ourdataset. 
Each training example is represented as $(I, q, r, a)$, where $I$ is the RGB observation, $q$ is the free-form instruction referring to the target object $o_t$, $r$ is the visually grounded reasoning chain describing the ancestor objects $\mathcal{A}(o_t)$, and $a$ is the final prediction corresponding to the top-level obstructors $\mathcal{F}(o_t)$. We concatenate $r$ and $a$ into a single output sequence $y=(y_1,\dots,y_{T_{r,a}})$.

The objective of SFT is to maximize the likelihood of generating both $r$ and $a$ conditioned on $(I,q)$. Using the standard autoregressive formulation, the training loss is

\begin{equation}
\mathcal{L}_{\text{SFT}}
= - \mathbb{E}_{(I,q,r,a)\sim\mathcal{D}}
\left[
\sum_{t=1}^{T_{r,a}}
\log p_{\Theta}(y_t \mid I, q, y_{<t})
\right]
\end{equation}
where $\mathcal{D}$ denotes the SFT training set, and $p_{\Theta}$ is the token-level conditional distribution produced by $f_{\Theta}$. This supervision encourages the model to output step-by-step obstruction chains that are physically grounded (\ie, each step corresponds to a contacting neighbor) and to predict the correct set of top-level obstructors. The best resulting model $f_{\Theta}^{\text{SFT}}$ with obstruction information serves as the initialization for the subsequent reinforcement fine-tuning (RFT) stage, ensuring stable optimization under task-specific rewards.

\subsection{RFT details}
\label{sec:add_rft}

Starting from the best SFT-trained model $f_{\Theta}^{\text{SFT}}$, we further optimize it with reinforcement fine-tuning using Group Relative Policy Optimization (GRPO). For each training prompt $(I, q)$ sampled from the RFT dataset $\mathcal{D}$, the current policy
$\pi_{\Theta}(\cdot \mid I, q)$ (induced by $f_{\Theta}^{\text{SFT}}$) generates $G$ candidate outputs
$\{y^{(g)}\}_{g=1}^{G}$, where each $y^{(g)} = (y^{(g)}_1,\dots,y^{(g)}_{T^{(g)}})$ is a full
\texttt{<think>}+\texttt{<answer>} sequence. Each generated sample receives a reward composed of format and task reward:
\[
r^{(g)} = \lambda_{\text{fmt}} \, r_{\text{fmt}}^{(g)} + \lambda_{\text{task}} \, r_{\text{task}}^{(g)},
\]
as defined in the main paper. GRPO uses group-wise relative advantages by subtracting the
within-group mean reward,
\[
\bar{r} = \frac{1}{G} \sum_{g=1}^{G} r^{(g)}, 
\qquad
A^{(g)} = r^{(g)} - \bar{r}.
\]

The RFT objective then minimizes the following loss:
{\small
\begin{align}
\mathcal{L}_{\text{RFT}}(\Theta)
&= - \mathbb{E}_{(I,q)\sim\mathcal{D}}
\Bigg[
\frac{1}{G} \sum_{g=1}^{G} A^{(g)} 
\sum_{t=1}^{T^{(g)}} 
\log \pi_{\Theta}\!\big(y^{(g)}_t \mid I, q, y^{(g)}_{<t}\big)
\Bigg]
\notag \\[1mm]
&\quad + \beta\,\mathrm{KL}\!\left(\pi_{\Theta}\,\|\,\pi_{\Theta}^{\text{SFT}}\right),
\end{align}
}
where $\pi_{\Theta_0}$ is the frozen SFT reference policy initialized from $f_{\Theta_0}$, and
$\beta$ controls the strength of KL regularization. The first term encourages generations with higher
relative rewards within each group, while the KL term keeps the updated policy close to the SFT
initialization, stabilizing optimization under task-specific rewards.

\section{Additional experimental details and analysis}
\label{sec:add_imp}
\subsection{Implementation details}
\noindent\textbf{\ourmethod.} Our method training is built on the Qwen official repo~\cite{bai2025qwen2} for SFT, and VLM-R1~\cite{shen2025vlm} for GRPO. For clarity, we list the training hyperparameters used in SFT and RFT in Tab.~\ref{tab:sft-hyperparams} and Tab.~\ref{tab:rft-hyperparams}, respectively.

    \begin{table}[h]
    \centering
    \caption{Supervised Fine-Tuning}
    \vspace{-2mm}
    \label{tab:sft-hyperparams}
    \small
    \begin{tabular}{l c}
    \toprule
    \textbf{Hyperparameter} & \textbf{Value} \\
    \midrule
    Model & Qwen2.5-VL-3B-Instruct \\
    Training epochs & 2 \\
    Learning rate & $1\mathrm{e}{-5}$ \\
    Batch size (per device) & 4 \\
    Gradient accumulation & 1 \\
    Precision & bf16 \\
    LR scheduler & Cosine \\
    Warmup ratio & 0.03 \\
    Gradient clipping & 1.0 \\
    Vision tuning & Frozen \\
    MM-MLP / LLM tuning & Enabled \\
    Max sequence length & 8192 \\
    Max image pixels & 12.8M \\
    DeepSpeed config & ZeRO-3 \\
    \bottomrule
    \end{tabular}
    \end{table}

    \begin{table}[h]
    \centering
    \caption{Reinforcement Fine-Tuning (GRPO)}
    \vspace{-2mm}
    \label{tab:rft-hyperparams}
    \small
    \begin{tabular}{l c}
    \toprule
    \textbf{Hyperparameter} & \textbf{Value} \\
    \midrule
    Training epochs & 1 \\
    Batch size (per device) & 8 \\
    Gradient accumulation & 2 \\
    Precision & bf16 \\
    Group size ($G$) & 4 \\
    Max completion length & 512 \\
    $\beta$ (KL penalty) & 0.04 \\
    Temperature & 1.0 \\
    Top-$p$ & 0.9 \\
    Attention backend & FlashAttention-2 \\
    Gradient checkpointing & Enabled \\
    Random seed & 42 \\
    DeepSpeed config & ZeRO-2 \\
    \bottomrule
    \end{tabular}
    \end{table}

\begin{figure*}[!t]
    \centering
    \includegraphics[width=0.88\textwidth]{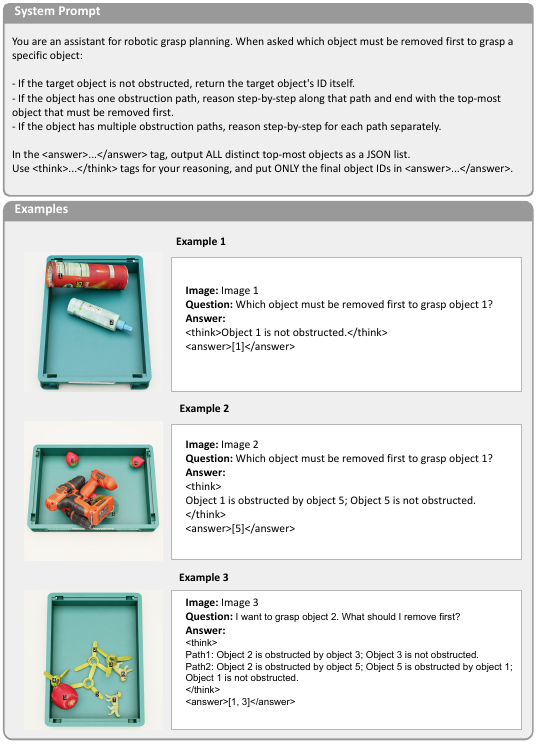}
    \caption{Synthetic set Oracle (SoM) examples for Qwen2.5-VL (ICL) and Gemini Robotics-ER 1.5 (ICL).}
    \label{fig:ICL1}
\end{figure*}

\begin{figure*}[t]
    \centering
    \includegraphics[width=0.88\textwidth]{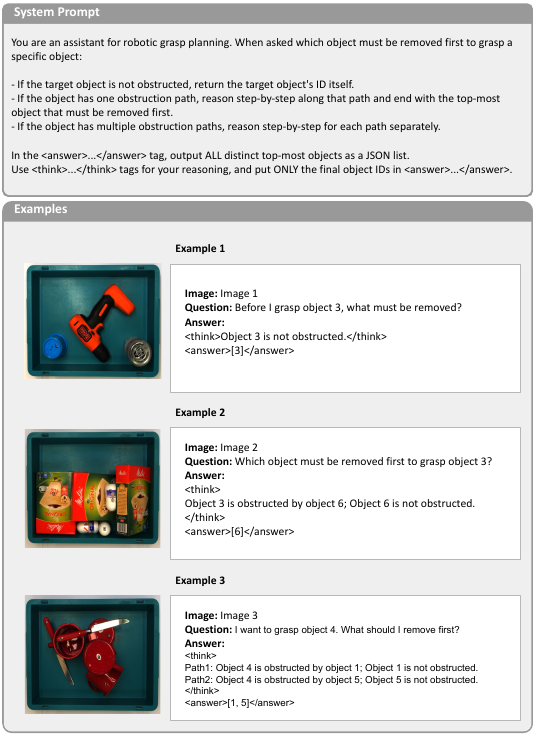}
    \caption{Real set Oracle (SoM) examples for Qwen2.5-VL (ICL) and Gemini Robotics-ER 1.5 (ICL).}
    \label{fig:ICL2}
\end{figure*}

\begin{figure*}[t]
    \centering
    \includegraphics[width=0.8\textwidth]{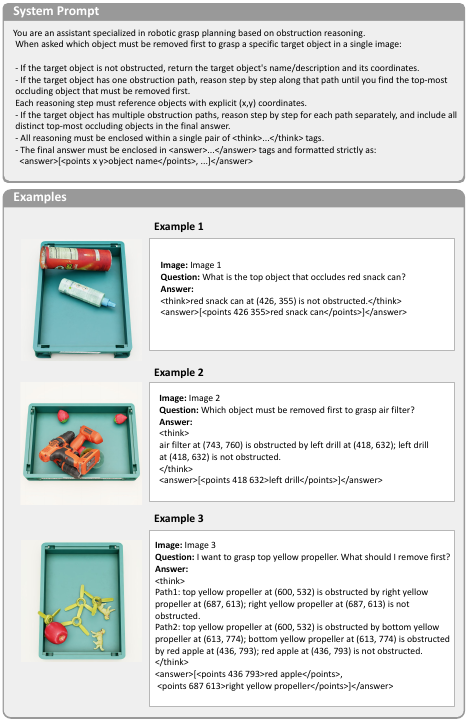}
    \caption{Synthetic set Natural Language Prompting examples for Qwen2.5-VL (ICL).}
    \label{fig:ICL3}
\end{figure*}

\begin{figure*}[t]
    \centering
    \includegraphics[width=0.8\textwidth]{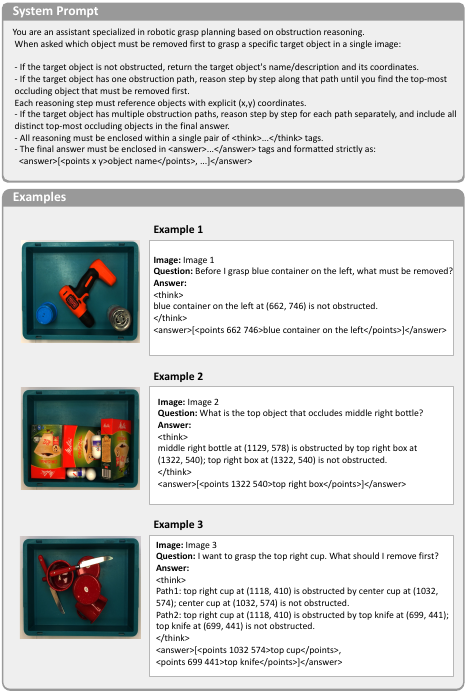}
    \caption{Real set Natural Language Prompting examples for Qwen2.5-VL (ICL).}
    \label{fig:ICL4}
\end{figure*}

\begin{figure*}[t]
    \centering
    \includegraphics[width=0.8\textwidth]{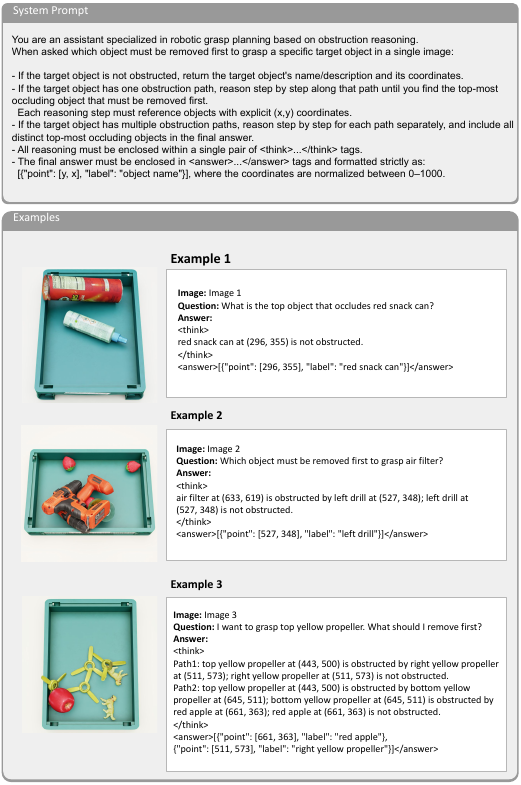}
    \caption{Synthetic set Natural Language Prompting examples for Gemini Robotics-ER 1.5 (ICL).}
    \label{fig:ICL5}
\end{figure*}

\begin{figure*}[t]
    \centering
    \includegraphics[width=0.8\textwidth]{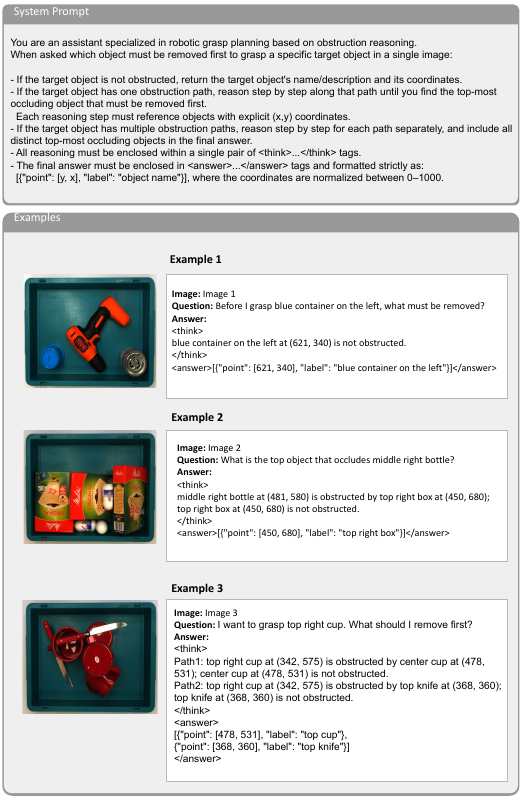}
    \caption{Real set Natural Language Prompting examples for Gemini Robotics-ER 1.5 (ICL).}
    \label{fig:ICL6}
\end{figure*}

\noindent\textbf{In Context Learning (ICL).}
We evaluate ICL for both Qwen2.5-VL and Gemini Robotics-ER 1.5 under four settings: Oracle (SoM) on Real scenes, NLP on Real scenes, Oracle (SoM) on Synthetic scenes, and NLP on Synthetic scenes. 
For each setting, three examples are randomly sampled from the corresponding subset (Synthetic or Real) to construct the few-shot ICL context.
In the Oracle (SoM) setting, both models receive exactly the same ICL examples (Fig.~\ref{fig:ICL1}, \ref{fig:ICL2}).
The only difference arises in the Natural Language Prompting setting: Qwen2.5-VL requires absolute pixel coordinates $(x, y)$ (Fig.~\ref{fig:ICL3}, \ref{fig:ICL4}), whereas Gemini Robotics-ER 1.5, based on their cookbook\footnote{https://ai.google.dev/gemini-api/docs/robotics-overview}, uses normalized coordinates $(y, x)$ within the range $[0,1000]$ (Fig.~\ref{fig:ICL5}, \ref{fig:ICL6}).

\subsection{Additional analysis}
\subsubsection*{Complete ablations for all variants}
\begin{figure}[t]
    \centering
    \includegraphics[width=0.75\linewidth]{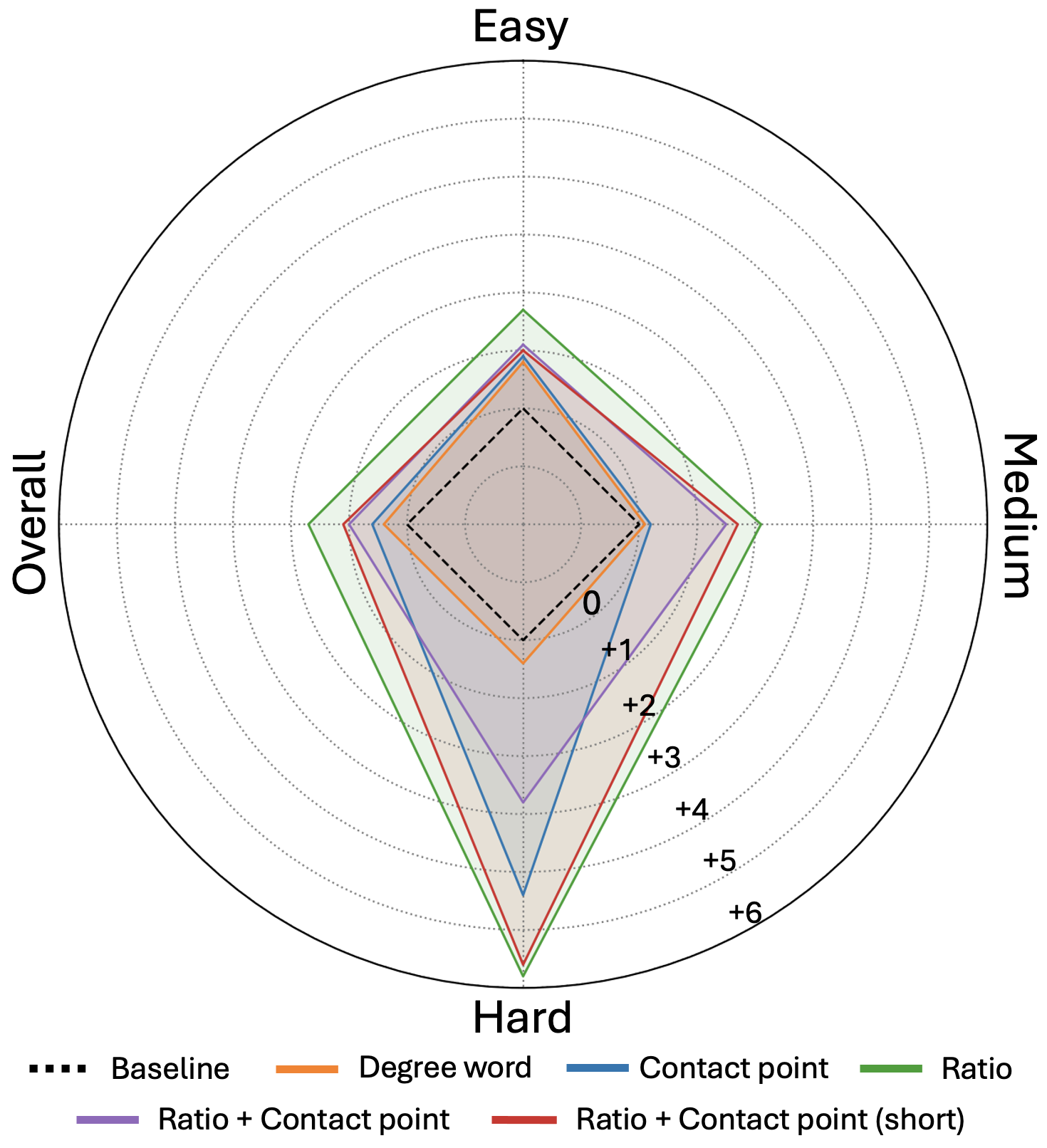}
    \vspace{-8pt}
    \caption{
        Radar chart illustrating the ablation study of obstruction information used during SFT. 
        We compare three difficulty levels (Easy, Medium, Hard) and Overall with different obstruction cues.
    }
    \label{fig:radar_obstruction_ablation}
    \vspace{-6pt}
\end{figure}

Tab.~\ref{tab:sft-occlusion-ablation-som} and Fig.~\ref{fig:radar_obstruction_ablation} present the numerical and visual results of incorporating different types of obstruction information during SFT across the Easy, Medium, Hard, and Overall setting. All obstruction cues consistently improve the baseline model's obstruction reasoning performance, and the improvement becomes more pronounced as the scene difficulty increases. Among them, obstruction ratio yields the strongest boost, achieving a 5.8\% increase in SR-F1 under the Hard setting, indicating that continuous ratio signals help learning complex obstruction reasoning.

\begin{figure*}[!t]
    \centering
    \includegraphics[width=0.8\textwidth]{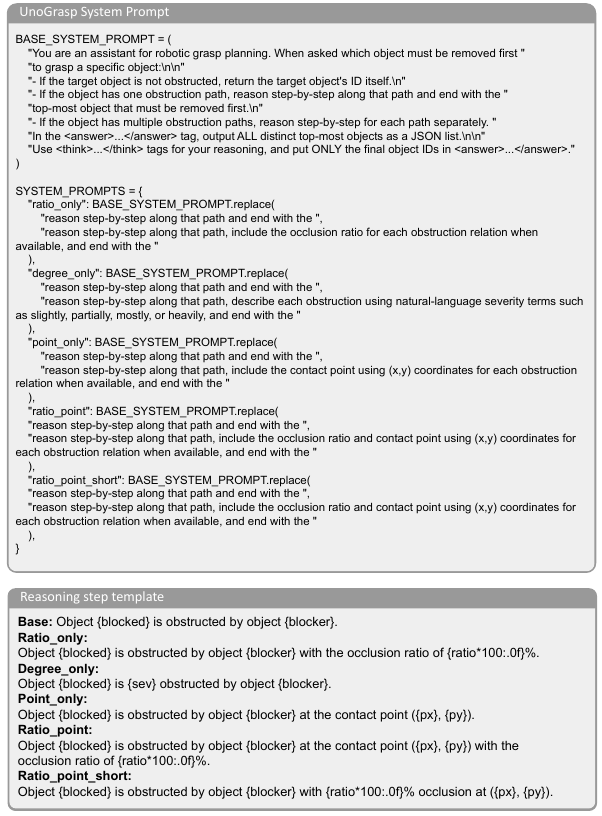}
    \caption{\ourmethod system prompt templates and corresponding reasoning formats for obstruction information.}
    \label{fig:ICL0}
\end{figure*}

We further combine the two most effective types of obstruction information, \ie, Obstruction Ratio and Contact Point, and evaluate both full-sentence and short-sentence templates (as shown in Fig.~\ref{fig:ICL0}). 
The short one achieves better performance generally, especially in Hard scenes. We hypothesize that this might be because shorter templates reduce unnecessary tokens, allowing the model to focus more on critical obstruction cues and thus produce higher quality reasoning in long and complex obstruction paths.
However, in the SoM setting, the benefits of joining Obstruction Ratio with Contact Point are not better, suggesting the need for better fusion strategies.

\begin{table*}[t]
\caption{Ablation on obstruction information in SFT (SOM variant). 
All SR-F1 and OR-F1 values are absolute (0--100 scale). }
\label{tab:sft-occlusion-ablation-som}
\tabcolsep 5pt
\resizebox{\textwidth}{!}{%
\begin{tabular}{l ccc ccc ccc ccc}
\toprule

& \multicolumn{3}{c}{\cellcolor{lightgreen}Easy} & \multicolumn{3}{c}{\cellcolor{lightyellow}Medium} & \multicolumn{3}{c}{\cellcolor{verylightpink}Hard} & \multicolumn{3}{c}{Overall}\\
\cmidrule(lr){2-4}\cmidrule(lr){5-7}\cmidrule(lr){8-10}\cmidrule(lr){11-13}
Method & SR-F1 $\uparrow$ & OR-F1 $\uparrow$ & MP\_NED $\downarrow$
       & SR-F1 $\uparrow$ & OR-F1 $\uparrow$ & MP\_NED $\downarrow$
       & SR-F1 $\uparrow$ & OR-F1 $\uparrow$ & MP\_NED $\downarrow$
       & SR-F1 $\uparrow$ & OR-F1 $\uparrow$ & MP\_NED $\downarrow$ \\
\midrule
Baseline (None)
& 80.1 & 79.2 & 0.125
& 65.0 & 58.5 & 0.396
& 44.3 & 45.7 & 0.543
& 74.7 & 71.9 & 0.220 \\

+ Contact Point
& 81.0 & 80.1 & 0.120
& 65.2 & 58.5 & 0.394
& 48.7 & 45.0 & 0.540
& 75.3 & 72.5 & 0.216 \\

+ degree word
& 80.9 & 79.8 & 0.125
& 65.1 & 59.2 & \textbf{0.388}
& 44.7 & 44.8 & 0.541
& 75.1 & 72.5 & 0.217 \\

+ Ratio
& \textbf{81.8} & \textbf{80.9} & \textbf{0.115}
& \textbf{67.1} & 59.4 & 0.389
& \textbf{50.1} & 46.9 & 0.525
& \textbf{76.4} & \textbf{73.3} & \textbf{0.210} \\

+ Ratio + Contact Point
& 81.2 & 80.3 & 0.118
& 66.5 & 58.7 & 0.389
& 47.1 & 47.0 & 0.529
& 75.7 & 72.7 & 0.212 \\

+ Ratio + Contact Point (Short)
& 81.1 & 80.3 & 0.116
& 66.7 & \textbf{59.5} & 0.392
& 49.9 & \textbf{49.2} & \textbf{0.517}
& 75.8 & 73.0 & 0.212 \\
\bottomrule
\end{tabular}
}
\end{table*}

\subsubsection*{More qualitative results.}
We present qualitative results on the UNOBench real set, comparing \ourmethod, Gemini Robotics-ER 1.5 (ICL), and Qwen2.5-VL (ICL). Notably, our method does not use any real-set data during training, whereas Gemini and Qwen are each provided with three real few-shot examples covering \emph{no-obstruction}, \emph{single-path}, and \emph{multi-path} ICL examples.

\begin{figure*}[t]
    \centering
    \includegraphics[width=0.82\textwidth]{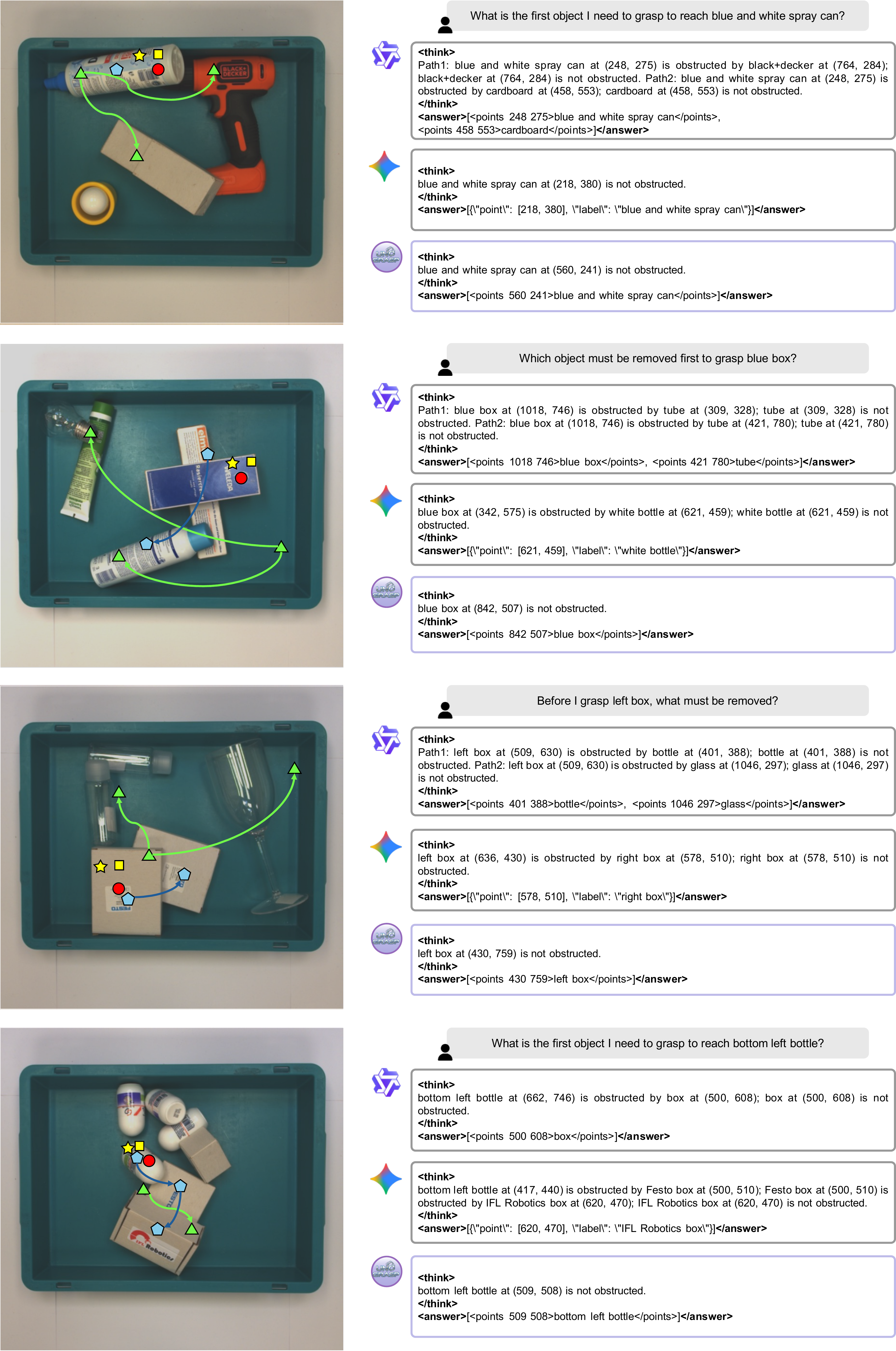}
    \caption{No-obstruction cases on the \ourdataset real set. \includegraphics[height=0.9em]{main/figures/icon/star_compressed.pdf} mark the target object, \includegraphics[height=0.9em]{main/figures/icon/square_compressed.pdf} the top obstructor, \includegraphics[height=0.9em]{main/figures/icon/red_compressed.pdf}~UNOGrasp, \includegraphics[height=0.9em]{main/figures/icon/blue_compressed.pdf}~Gemini Robotics-ER~1.5, 
and \includegraphics[height=0.9em]{main/figures/icon/green_compressed.pdf}~Qwen2.5-VL (ICL) predictions with their reasoning traces.}
    \label{fig:real1}
\end{figure*}

Fig.~\ref{fig:real1} shows no-obstruction examples. Even though the ICL context explicitly includes a similar case, both Gemini Robotics-ER 1.5 and Qwen2.5-VL hallucinate the obstruction among objects that are far apart. In contrast, with grounded obstruction information, \ourmethod effectively suppresses such hallucinations. Additionally, we observe that for object pairs involved in an obstruction relation, Gemini Robotics-ER 1.5 and Qwen2.5-VL tend to regard the queried target as the \emph{object being obstructed}, even when the target is actually the one \emph{doing the obstructing}.

\begin{figure*}[t]
    \centering
    \includegraphics[width=0.82\textwidth]{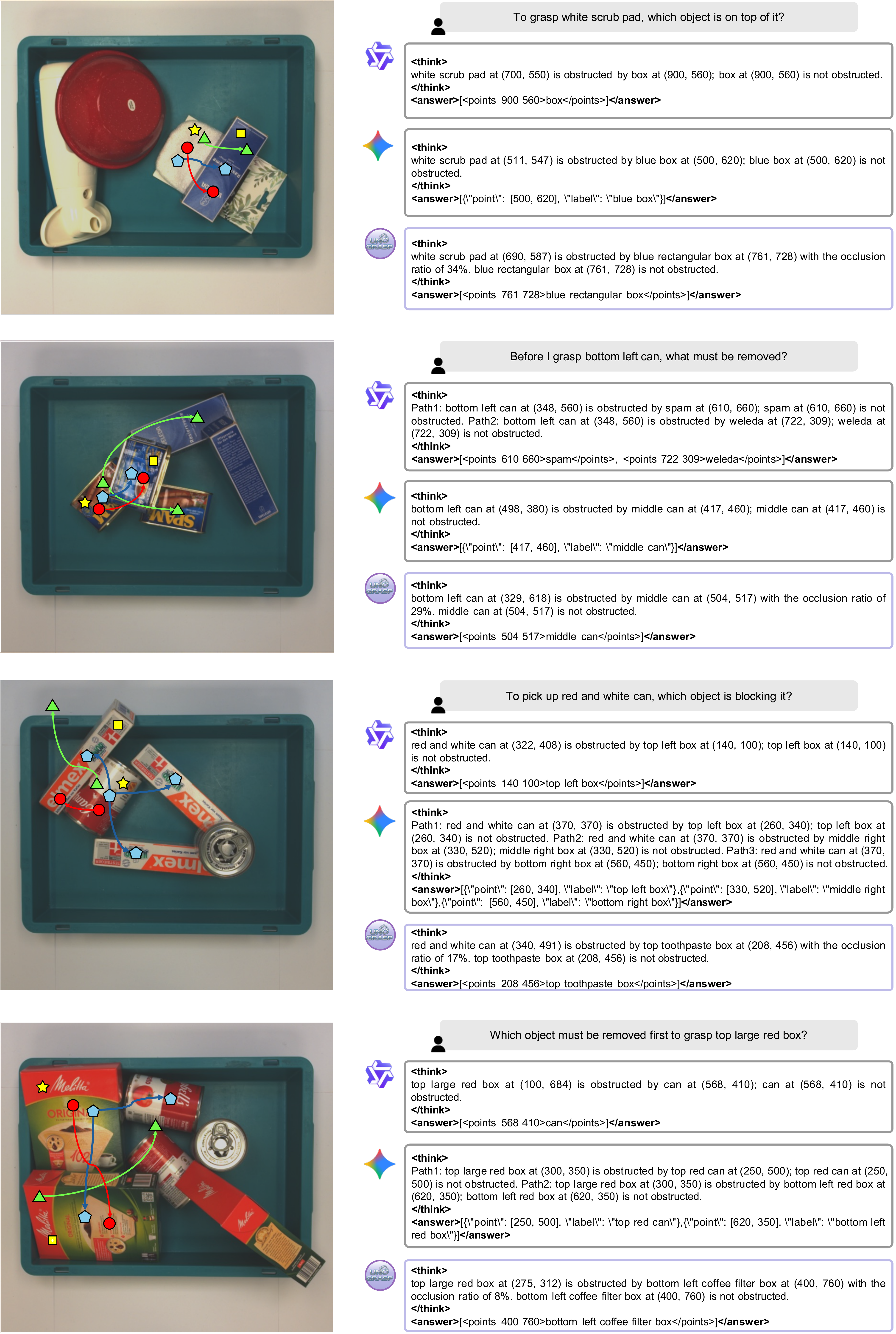}
    \caption{Easy cases on the \ourdataset real set. \includegraphics[height=0.9em]{main/figures/icon/star_compressed.pdf} mark the target object, \includegraphics[height=0.9em]{main/figures/icon/square_compressed.pdf} the top obstructor, \includegraphics[height=0.9em]{main/figures/icon/red_compressed.pdf}~UNOGrasp, \includegraphics[height=0.9em]{main/figures/icon/blue_compressed.pdf}~Gemini Robotics-ER~1.5, 
and \includegraphics[height=0.9em]{main/figures/icon/green_compressed.pdf}~Qwen2.5-VL (ICL) predictions with their reasoning traces.}
    \label{fig:real2}
\end{figure*}

Fig.~\ref{fig:real2} presents Easy examples.\ourmethod and Gemini Robotics-ER 1.5 both produce mostly correct predictions in this case, but Gemini is still prone to misidentifying spatially close objects as obstructors.
On the other hand, Qwen2.5-VL exhibits more severe hallucinations: the predicted object names and coordinates are frequently misaligned, and the reasoning in \texttt{<think>} sometimes directly contradicts the final \texttt{<answer>} provided.

\begin{figure*}[t]
    \centering
    \includegraphics[width=0.82\textwidth]{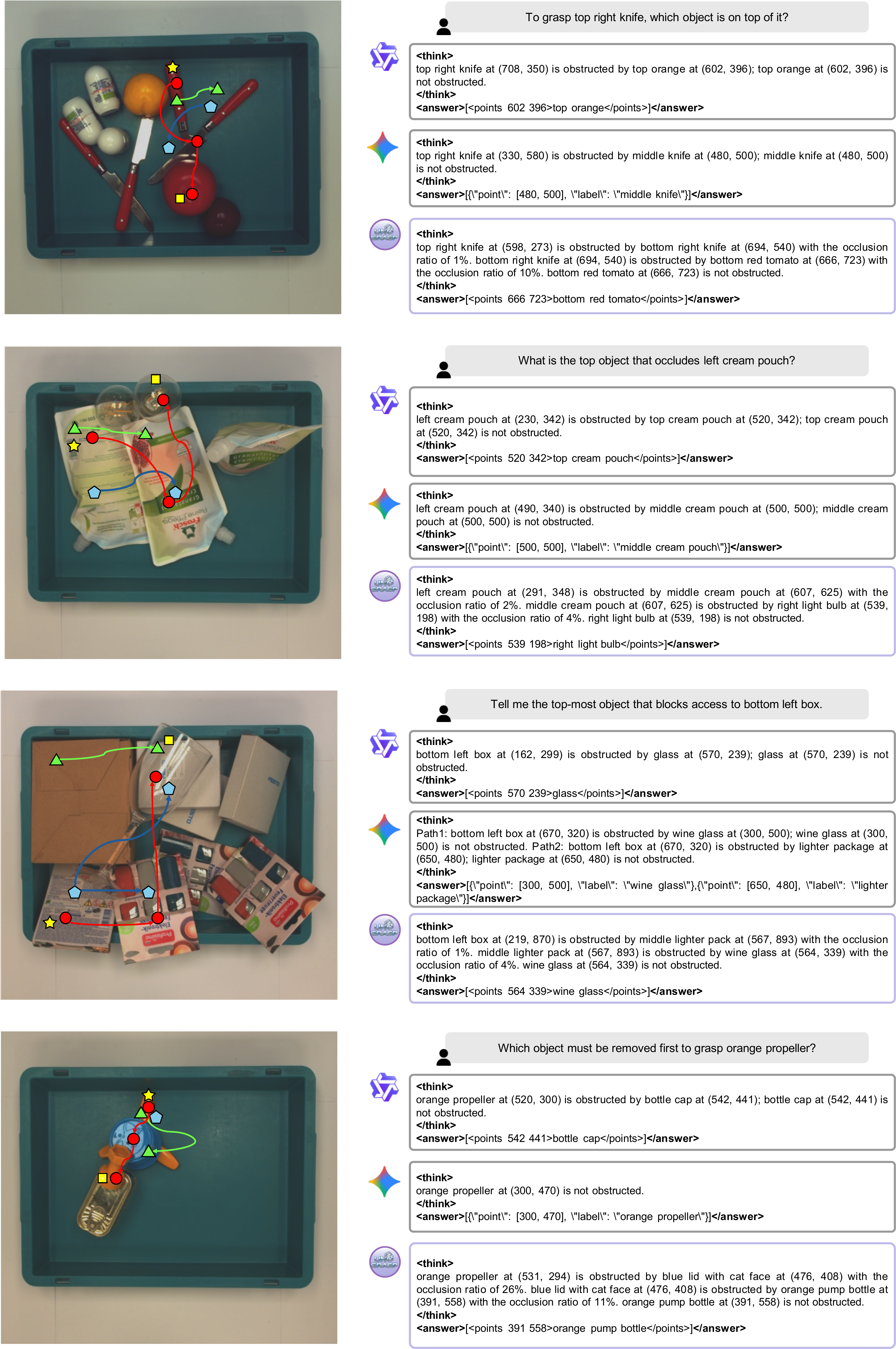}
    \caption{Medium cases on the \ourdataset real set. \includegraphics[height=0.9em]{main/figures/icon/star_compressed.pdf} mark the target object, \includegraphics[height=0.9em]{main/figures/icon/square_compressed.pdf} the top obstructor, \includegraphics[height=0.9em]{main/figures/icon/red_compressed.pdf}~UNOGrasp, \includegraphics[height=0.9em]{main/figures/icon/blue_compressed.pdf}~Gemini Robotics-ER~1.5, 
and \includegraphics[height=0.9em]{main/figures/icon/green_compressed.pdf}~Qwen2.5-VL (ICL) predictions with their reasoning traces.}
    \label{fig:real3}
\end{figure*}

Fig.~\ref{fig:real3} illustrates more complex Medium cases with multiple obstruction paths or deeper reasoning chains. Here, \ourmethod typically produces complete and coherent reasoning, tracing each path up to the correct top-level obstructor. Gemini Robotics-ER 1.5 tends to terminate the reasoning prematurely before reaching the true top obstructor, resulting in incorrect predictions.


\subsubsection*{Failure-case visualizations.}
\begin{figure*}[t]
    \centering
    \includegraphics[width=0.82\textwidth]{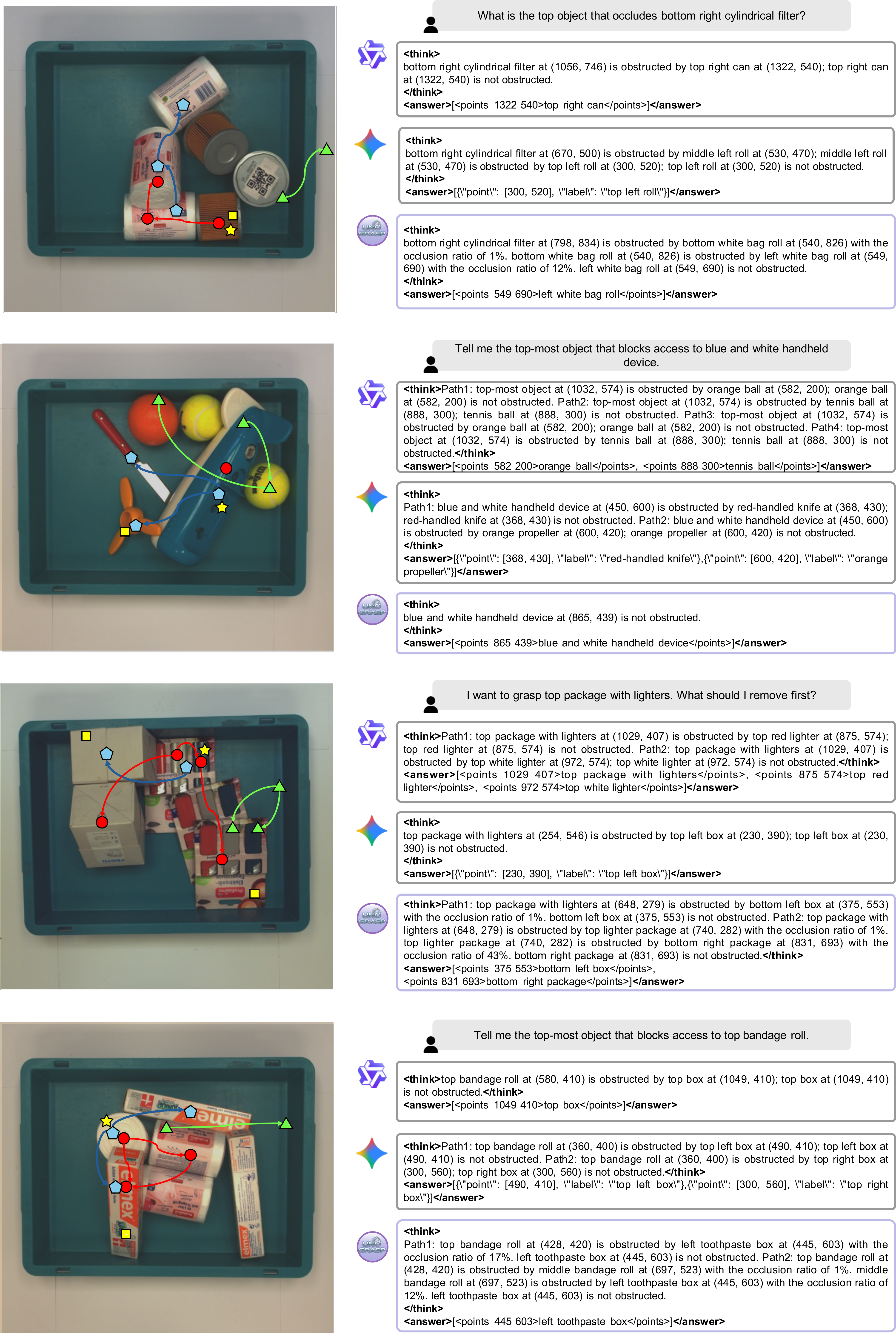}
    \caption{Real set failed cases. \includegraphics[height=0.9em]{main/figures/icon/star_compressed.pdf} mark the target object, \includegraphics[height=0.9em]{main/figures/icon/square_compressed.pdf} the top obstructor, \includegraphics[height=0.9em]{main/figures/icon/red_compressed.pdf}~UNOGrasp, \includegraphics[height=0.9em]{main/figures/icon/blue_compressed.pdf}~Gemini Robotics-ER~1.5, 
and \includegraphics[height=0.9em]{main/figures/icon/green_compressed.pdf}~Qwen2.5-VL (ICL) predictions with their reasoning traces.}
    \label{fig:real4}
\end{figure*}

Fig.~\ref{fig:real4} summarizes several typical failure cases of \ourmethod. 
The first category arises when objects are in physical contact but do not form an actual obstruction. In these cases, the model incorrectly predicts an obstruction relationship. Even when the model outputs a low occlusion ratio (\eg, 1\%), indicating minimal obstruction, it still struggles to distinguish mere contact from true obstruction.

The second category occurs in scenes containing multiple objects with similar features (\eg, shapes and colors). When such objects touch and form obstructions, the model tend to fail at both the detection stage and the subsequent obstruction reasoning stage, ultimately leading to incorrect identification of the obstruction paths.

\section{Real-Robot Experiments}
\label{sec:add_robot}

We conducted extensive real-world experiments comparing our approach against state-of-the-art VLM baselines.
The experiments were designed to test obstruction reasoning in cluttered environments, specifically evaluating the generalization capability of VLMs in unseen real-world setups. We show the models' outputs and the subsequent robotic executions under the Easy, Medium and Hard scenarios in the demonstration video. Here below, we describe in details the robotic setup and the experimental procedure.

\subsection{Robotic Setup}

\begin{figure*}[t]
    \centering
    \includegraphics[width=0.85\textwidth]{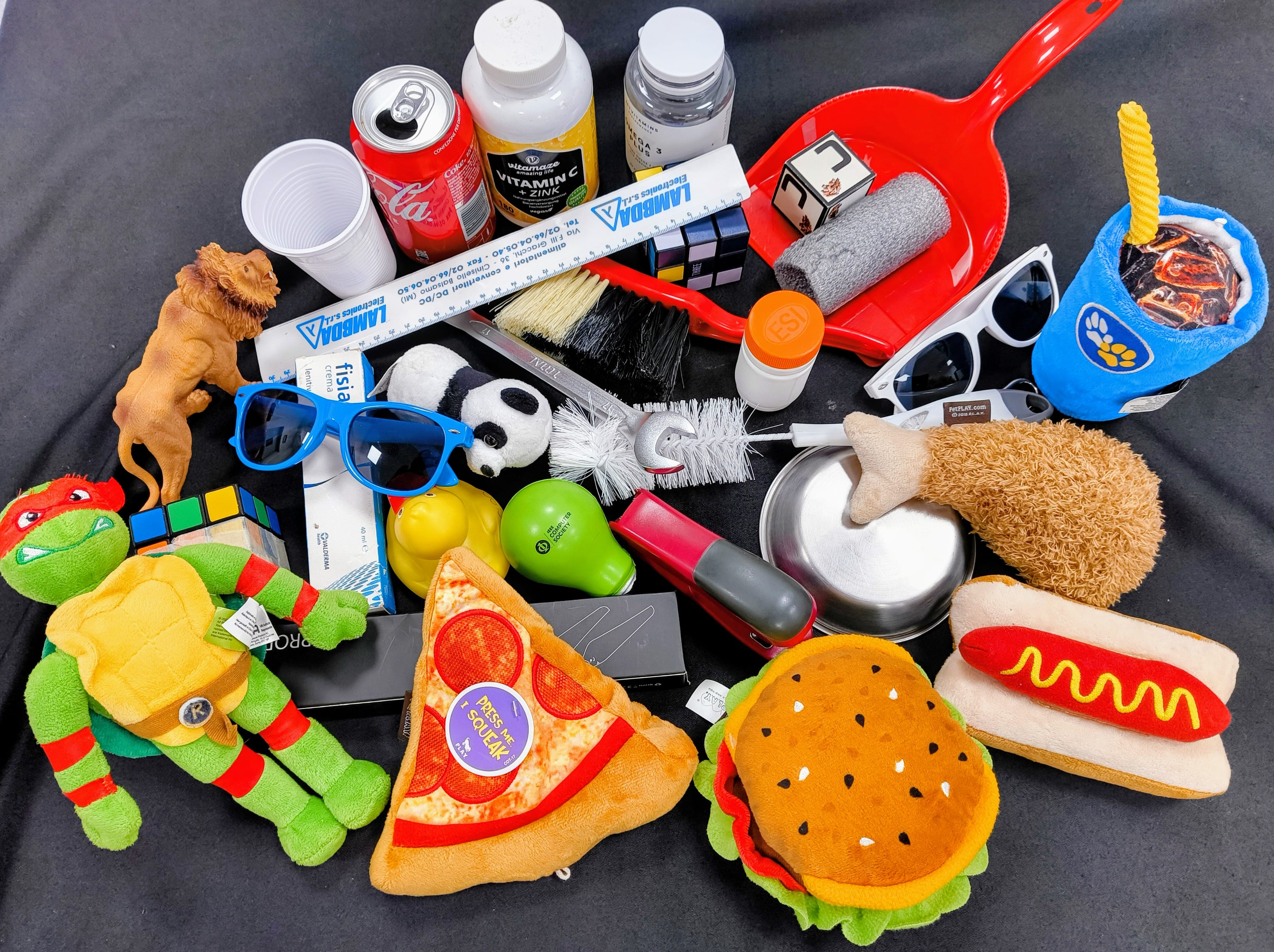}
    \caption{
    \textbf{Object set used in Real-Robot experiments.}
    The set comprises 25 distinct household objects varying in shape, size, texture, and deformability. These objects were arranged to create 30 unique scenarios ranging from Easy (shallow piles) to Hard (deep occlusion chains).
    }
    \label{fig:experiment_objects}
\end{figure*}

The experimental platform consists of a Universal Robots UR5e 6-DoF manipulator equipped with a Robotiq 2F-85 parallel-jaw gripper.
The observation is provided by a Stereolabs ZED 2 stereo camera mounted in a top-down configuration, approximately 0.8\,m above the workspace.
The workspace contains a bin populated with a diverse set of rigid objects (see Fig.~\ref{fig:experiment_objects}).

To ensure a fair comparison, the experimental framework integrates every VLM reasoning agent with an identical grasp generation backend.
We utilize GraspNet~\cite{Gilles2024MetaGraspNetV2} to generate a dense set of candidate 6-DoF grasp poses ($SE(3)$) globally across the scene, and GroundedSAM~\cite{ren2024groundedSAM} to filter and select the specific poses associated with the object mask predicted by the VLM.
This isolation ensures that performance differences are attributable solely to the reasoning capabilities of the tested VLMs, \ie \ourmethod, Gemini Robotics-ER 1.5, and Qwen2.5-VL.

\subsection{Experimental Procedure}
\label{subsec:procedure}
The experiments follow a step-wise protocol to ensure that all models are evaluated on identical scene configurations at each individual time step.

\noindent\textbf{Initialization.}
For each scenario, the operator arranges the objects in the bin and initializes the system with three parameters:
(1) the \textit{User Prompt} (\eg, ``grasp the red block''), which remains constant across all time steps;
(2) the \textit{Number of Objects} in the scene; and
(3) the \textit{Maximum Path Length} ($k_{max}$), defining the expected number of removal steps.

\noindent\textbf{Execution Loop.}
The procedure operates in discrete time steps. At each step $t$, all VLMs will perform obstruction reasoning and predict the next move given same the physical state of the bin. The cycle is as follows:

\begin{enumerate}
    \item \textbf{Observation:} The ZED 2 camera captures RGB-D data, which is normalized and cropped to the bin area.
    \item \textbf{Obstruction Reasoning:} The VLM processes the visual data and the prompt to reason on the obstruction and predict the next object to be removed.
    \item \textbf{Operator Feedback (Validation):} The human operator verifies if the predicted object by the VLM is a valid object that is reachable at the current step. If the prediction is not a top-level obstructor or the grasping point is on the bin itself, the robotic execution will be aborted. Crucially, if a VLM fails this validation, it will be excluded from the subsequent steps and marked as a failure.
    \item \textbf{Execution:} If the prediction is validated, the robot executes the grasp using the integrated planner.
    \item \textbf{Operator Tidy Up (Reset):} After the robot complete its execution, the operator manually restores the scene. Note that it is likely during the grasping operations, other objects are displaced accidentally. To ensure fair evaluation of VLMs on receiving the same scene observations at each time step, the operator resets the objects' arrangement to their original setup.
\end{enumerate}

Once all VLMs have been evaluated for step $t$, the object identified as the correct obstruction is removed from the bin to advance the physical state to step $t+1$.
This loop repeats until the target object is grasped or all methods fail.

\end{document}